\documentclass{article} 
\usepackage{iclr2025_conference,times}

\usepackage{amsmath,amsfonts,bm}

\def\eqref#1{equation~\ref{#1}}

\def\1{\bm{1}}

\DeclareMathAlphabet{\mathsfit}{\encodingdefault}{\sfdefault}{m}{sl}
\SetMathAlphabet{\mathsfit}{bold}{\encodingdefault}{\sfdefault}{bx}{n}

\usepackage[utf8]{inputenc} 
\usepackage[T1]{fontenc}    

\usepackage{hyperref}
\usepackage{url}        
\usepackage{booktabs}       
\usepackage{amsfonts}       
\usepackage{nicefrac}       
\usepackage{microtype}      
\usepackage{xcolor}         
\usepackage{dsfont}
\usepackage{float}
\usepackage{multirow}
\usepackage{subcaption}
\usepackage{graphicx}
\usepackage{array}
\usepackage{listings}
\usepackage{enumitem,kantlipsum}

\newcommand{\method}[1]{BenchBuilder}
\newcommand{\dataset}[1]{Arena-Hard-Auto}
\newcommand{\wildset}[1]{Wild-Hard-Auto}
\newcommand{\llamaset}[1]{Llama-Hard-Auto}

\usepackage{tcolorbox}
\tcbuselibrary{most}
\newtcolorbox{userinput}[1]{
    enhanced,
    drop shadow=black!5!white,
    left=4mm,
    right=4mm,
    top=3mm,
    bottom=3mm,
    boxsep=0mm,
    rounded corners,
    title=#1,
    fontupper=\linespread{1.1}\scriptsize\fontfamily{lmr}\selectfont
    }

\newtcolorbox{prompt}[1]{
    enhanced,
    drop shadow=black!5!white,
    left=4mm,
    right=4mm,
    top=3mm,
    bottom=3mm,
    boxsep=0mm,
    rounded corners,
    title=#1,
    fontupper=\linespread{1.1}\fontfamily{lmr}\selectfont
    }

\newtcolorbox{example}{
    colback=yellow!10,
    colframe=yellow!40,
    fonttitle=\bfseries,
    coltitle=black,
    colbacktitle=yellow!60,
    enhanced,
    drop shadow=black!5!white,
    left=2mm,
    right=2mm,
    top=3mm,
    bottom=3mm,
    boxsep=0mm,
    sharp corners,
    float
    }

\iclrfinalcopy

\title{From Crowdsourced Data to High-Quality Benchmarks: Arena-Hard and BenchBuilder Pipeline}

\author{%
Tianle Li\thanks{Equal contribution. Correspondence to: Tianle Li (tianleli@berkeley.edu), Wei-Lin Chiang (weichiang@berkeley.edu)} \quad Wei-Lin Chiang$^*$ \quad Evan Frick \quad Lisa Dunlap \quad Tianhao Wu \\
\textbf{Banghua Zhu} \quad \textbf{Joseph E. González} \quad \textbf{Ion Stoica} \\ \\
UC Berkeley \\
}

\begin{document}

\maketitle

\begin{abstract}
The rapid evolution of Large Language Models (LLMs) has outpaced the development of model evaluation, highlighting the need for continuous curation of new, challenging benchmarks. However, manual curation of high-quality, human-aligned benchmarks is expensive and time-consuming. To address this, we introduce \method{}, an automated pipeline that leverages LLMs to curate high-quality, open-ended prompts from large, crowd-sourced datasets, enabling continuous benchmark updates without human in the loop. We apply \method{} to datasets such as Chatbot Arena and WildChat-1M, extracting challenging prompts and utilizing LLM-as-a-Judge for automatic model evaluation. To validate benchmark quality, we propose new metrics to measure a benchmark’s alignment with human preferences and ability to separate models. We release \dataset{}, a benchmark consisting 500 challenging prompts curated by \method{}. \dataset{} provides 3x higher separation of model performances compared to MT-Bench and achieves 98.6\% correlation with human preference rankings, all at a cost of \$20. Our work sets a new framework for the scalable curation of automated benchmarks from extensive data.
\end{abstract}

\section{Introduction}
The proliferation of Large Language Models (LLMs) has spurred advancements as models expand their capabilities by training on increasingly vast and diverse datasets. Traditional static benchmarks~\citep{wang2019glue, rajpurkar2016squad, bowman-etal-2015-large, dolan2005automatically, bos-markert-2005-recognising, mmlu} are quickly becoming saturated and struggle to differentiate state-of-the-art models.

To address these limitations, recent benchmarks like GPQA~\citep{rein2023gpqa} source high-quality and challenging prompts from domain experts. Although these efforts have produced challenging evaluation sets, they come at a steep price---GPQA, for instance, cost over \$120,000 to curate its 500 multiple-choice questions~\citep{gpqacostblog2024}. The reliance on manual curation makes such benchmarks difficult to produce. Moreover, their static nature is susceptible to test-set leakage and overfitting as models are trained on similar datasets. This necessitates the continuous development of new benchmarks, exacerbating the cost and labor of manual curation. Further, many of these benchmarks rely on close-ended tasks that fail to capture the open-ended nature of real-world interactions, undermining their cost-effectiveness for evaluating alignment to user preference.

\begin{figure}[t]
\centering
\includegraphics[width=0.75\textwidth]{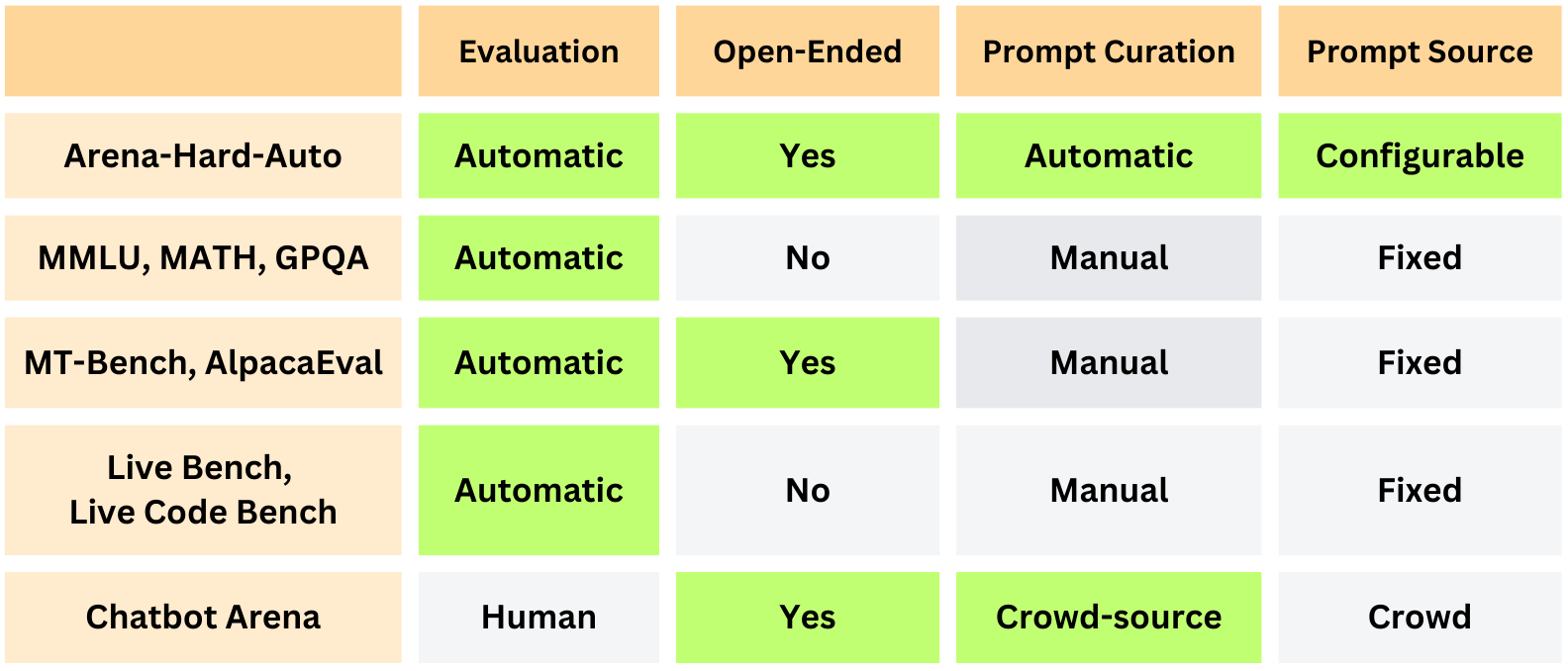}
\caption{Classification of LLM benchmarks: we categorize benchmarks on how the evaluation can be done, whether the evaluated tasks are ground-truth or open-ended, how are the prompts curated, and whether the developer can control the source for the prompts.}
\label{fig:benchmark_classification}
\vspace{-1em}
\end{figure}

An alternative approach without manual curation involves crowdsourcing prompts through live evaluation platforms such as Chatbot Arena~\citep{chiang2024chatbot}. These platforms test models against a continuous stream of fresh, open-ended queries and user feedback. However, real-time human evaluation is both expensive and time-consuming, rendering these platforms infeasible for frequent evaluations by model developers. Moreover, while the crowd-sourced prompts represent real-world and open-ended tasks, their quality varies in difficulty and cannot be converted to challenging benchmarks without careful data filtering.  

In light of these open challenges, there is a pressing need for an automated pipeline which can curate high-quality prompts dynamically at scale. In this paper, we introduce \method{}, an automated benchmark curation system designed to address these gaps. \method{} leverages LLMs to curate, filter, and validate prompts based on seven indicators of high-quality prompts, such as specificity and domain knowledge, creating a pipeline that can continuously curate benchmarks alongside model development. 

We apply \method{} to crowd-sourced datasets, both Chatbot Arena~\citep{chiang2024chatbot} and WildChat-1M~\citep{zhao2024wildchat}, demonstrating that it can robustly generate high-quality benchmarks that differentiate models. The resulting benchmark, \dataset{}, employs LLM judges~\citep{zheng2023judging, alpaca_eval} to estimate human preferences against a baseline model, making the entire process—from prompt curation to evaluation—fully automated. We also address potential biases in LLM-based evaluations and propose solutions to mitigate them. To assess benchmark quality, we introduce new metrics that measure a benchmark’s ability to confidently separate models and align with human preferences. When compared to leading benchmarks such as AlpacaEval LC~\citep{dubois2024length} and MT-Bench~\citep{zheng2023judging}, \dataset{} achieves stronger model separability, tighter confidence intervals, and achieve 98.6\% correlation with Chatbot Arena rankings, making it a fast, reliable predictor of downstream model performance.

To summarize, our works makes the following contributions:
\begin{enumerate}
    \item We propose a novel data curation pipeline, \method{}, to automatically construct high-quality benchmarks from crowdsourced data.
    \item We propose metrics to capture desired properties in an LLM benchmark, and validate that \dataset{} achieves higher model separation and alignment to human preference than existing benchmarks. 
    \item We open-source both \method{} pipeline and \dataset{} benchmark\footnote{Our code is available at: \url{https://github.com/lmarena/arena-hard-auto}}.
\end{enumerate}

\section{Related works}
\label{sec:related}


\textbf{LLM benchmarks.} We briefly review widely used LLM benchmarks. Most existing benchmarks are static and ground-truth-based (e.g., multi-choice question answering). They cover a wide range of domains, including math, science, coding, and reasoning. Common ones include MMLU~\citep{mmlu}, MATH~\citep{hendrycksmath2021}, GSM-8K~\citep{cobbe2021gsm}, 
HumanEval~\citep{chen2021evaluating},
DROP~\citep{dua2019drop},
BigBench~\citep{srivastava2023beyond}, HellaSwag~\citep{zellers2019hellaswag}, AGIEval~\citep{zhong2023agieval}, GPQA~\citep{rein2023gpqa}, as well as comprehensive collection like HELM~\citep{liang2022holistic}.
Many have considered task-based evaluation such as IFEval~\citep{zhou2023instruction}, SWE-Bench~\citep{jimenez2024swebench}, BigCodeBench~\citep{zhuo2024bigcodebenchbenchmarkingcodegeneration} or AgentBench~\citep{liu2023agentbench}. As LLMs become widely adopted in open-ended scenarios involving interaction with humans (e.g., chatbot), many have considered human evaluation using domain experts or crowd raters such as Amazon Mechanical Turk~\citep{karpinska-etal-2021-perils,wang-etal-2023-self-instruct} to examine models' response quality.
As an alternative to human labeling, previous work has shown that LLM-as-a-judge can be effective human preference proxies (e.g., AlpacaFarm~\citep{dubois2023alpacafarm}, MT-bench~\citep{llmjudge}, AlpacaEval~\citep{alpaca_eval}, WildBench~\citep{lin2024wildbench}).

\textbf{Benchmark leakage.} A fundamental limitation of static benchmarks is the potential risk of test set leakage (i.e., contamination). Existing works~\citep{carlini2021extracting,sainz-etal-2023-nlp,yang2023rethinking, reid2024gemini} have suggested a growing risk of contamination, which undermines the reliability of benchmarks over time, motivating the need for benchmarks that are more frequently updated.

\textbf{Live benchmarks.} DynaBench~\citep{dynabench2021} identifies these
challenges and recommends creating living and continuously evolving benchmarks. Recent works LiveBench~\citep{white2024livebench}, LiveCodeBench~\citep{jain2024livecodebench}, MixedEval~\citep{ni2024mixeval}, R2E~\citep{jain2024r2e}, as well as the community based live evaluation, Chatbot Arena~\citep{chiang2024chatbot}. However, none of these focus on developing a pipeline for automatic benchmark curation to enable automatic evaluation on open-ended tasks.

\section{How do you measure benchmarks?} 
\label{sec:measurement}

\newcommand{\pr}{\mathrm{p}}
\newcommand{\out}{\mathrm{o}}
\newcommand{\score}{f}

We outline two key properties that the benchmark aiming to approximate human preference should possess to provide meaningful comparisons between models:
\begin{enumerate}[leftmargin=*,itemsep=0pt]
    \item \textbf{Separability:} the benchmark should separate models with high confidence.
    \item \textbf{Alignment with Human Preference:} the benchmark should agree with human preference.
\end{enumerate}

While previous works have focused on alignment, separability is also a crucial consideration when comparing models of similar quality (e.g., different checkpoints from the same training run). However, achieving high-confidence separability is challenging due to limitations in prompt design and inherent variances in LLM evaluations. Overly simplistic prompts fail to distinguish between models, while the randomness in human and LLM judgments leads to inconsistent predictions. As a result, it is often difficult to confidently determine if a model's apparent performance reflects a genuine difference in capability or merely noisy observations, highlighting a need for methods to verify whether a benchmark can reliably separate similar models.

Statistical measures like Pearson~\citep{pearson} and Spearman Correlations~\citep{spearman1961proof}, commonly used in benchmarks such as AlpacaEval~\citep{alpaca_eval} to measure correlation to human preference ranking, may fail to adequately address model separability and ranking instability. In addition, these measures only provide a coarse signal of ranking correlation without quantifying the magnitude of performance differences between model pairs.

To address these shortcomings, we develop three novel metrics: \textit{Separability with Confidence}, \textit{Agreement with Confidence}, and \textit{Pair Rank Brier Score}.

\textbf{Separability with Confidence} 
quantifies the benchmark's confidence by measuring its consistency in predicting the winner of a model pair across random seeds through bootstrapping. This is done by calculating the percentage of model pairs that have non-overlapping confidence intervals of their benchmark scores. A higher percentage indicates that the benchmark is more confident in distinguishing between the performance of different models, as the confidence intervals of their scores do not overlap.

\textbf{Agreement with Confidence Interval} measures how well benchmarks A and B confidently distinguish between two models with the same ordering. Given models $\pi_1, \pi_2$, we assign scores based on:
\begin{enumerate}[leftmargin=*,itemsep=0pt]
\item If both benchmarks confidently separate $\pi_1$, $\pi_2$, a score of 1 is assigned if their preference agree, and -1 if they disagree.
\item If either A or B cannot separate $\pi_1, \pi_2$ with confidence, we assign a score of 0. 
\end{enumerate}

The final agreement score is the average across all unique model pairs. A score of 1 implies perfect agreement with full confidence, while a score of -1 indicates complete disagreement.

\textbf{Pair Rank Brier Score} further assesses an LLM benchmark's capability to predict the ranking of a pair of competing models by rewarding confidence in correct predictions while penalizing confidence when incorrect. Consider two models $\pi_1 > \pi_2$ with disparate quality. Although two benchmarks A and B predict the same ranking $\pi_1 > \pi_2$, they predict $P(\pi_1 > \pi_2)$ as $.60$ and $.90$, respectively (undetectable by Spearman correlation). These benchmarks would result in very different Brier scores, reflecting their ability to quantify the magnitude of performance difference between the models. If both benchmarks give the wrong prediction of the winner, we prefer the benchmark with a less confident prediction. In other words, Brier score weighs a benchmark's accuracy and its ability to quantify the appropriate level of uncertainty in its predictions. Background on Pair Rank Brier Score can be found in Appendix~\ref{appendix:A.1}.

While no single metric is intended to be individually sufficient, we claim that together, these metrics offer a robust framework for assessing benchmark performance, balancing the need for clear differentiation with alignment to human preferences.

\section{The \method{} Pipeline and \dataset{} Dataset}
\label{sec:method}
\subsection{\method{}}
\label{sec:dataset_creation}

\begin{figure}[t]
\centering
\includegraphics[width=1\textwidth]{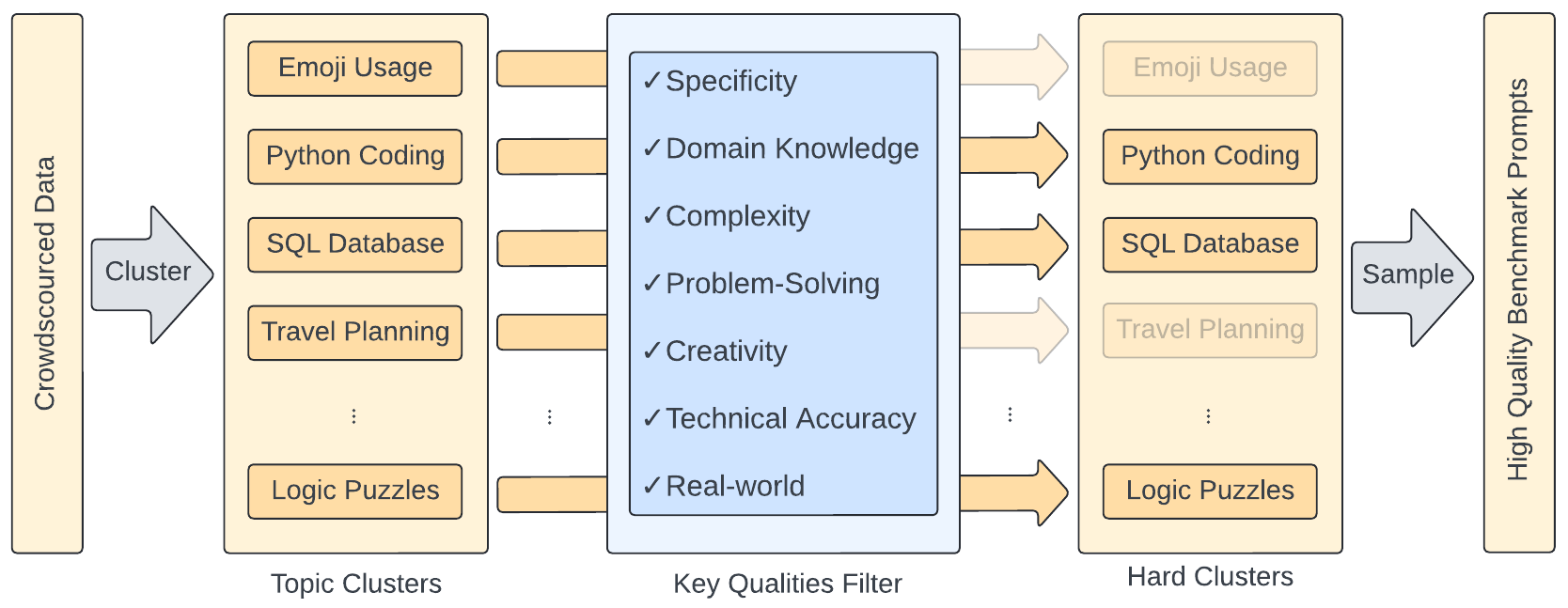}
\caption{\textbf{\method{} Pipeline.} Starting with a live data source of crowdsourced user prompts, we first cluster their embeddings to form topic clusters. An LLM annotator then assigns quality scores based on the required skills. Clusters with low quality scores are filtered out, and we sample from the remaining high-quality clusters to create a diverse and challenging dataset of benchmark prompts.}
\label{fig:method}
\end{figure}

The core idea behind how \method{} extract high-quality user queries from vast datasets is simple: each prompt is evaluated using a quality score, and prompts with high scores are sampled evenly across diverse topics. Figure~\ref{fig:method} illustrates our data creation pipeline.

To identify high-quality prompts, we define seven key qualities that capture the skills necessary to effectively address a query, such as specificity, domain expertise, and creativity (shown in Figure~\ref{fig:method}). An LLM-based annotator automatically scores each prompt by assessing how many of these qualities are present, producing a ``quality score''. Detailed instructions for these quality assessments are provided in Section~\ref{sec:prompts}.

To ensure our filtered prompts span a wide range of tasks, we leverage a topic modeling approach using BERTopic. We first encode each prompt using OpenAI’s embedding model, text-embedding-3-small~\citep{openai2024embedding}, reduce dimensions with UMAP, and apply a hierarchical-based clustering algorithm (HDBSCAN). This process generates distinct topic clusters. Each topic is then summarized and named using an LLM. 

Since some topic clusters predominantly contain trivial or poorly defined prompts (e.g., "hi"), we retain only the clusters with high average quality scores and sample prompts evenly across these selected clusters. The resulting dataset consists of mostly well-defined, technical problem-solving queries as required in the above key criteria. Dataset statistics and further details on our filtering and sampling strategy are provided in the following section.

\begin{prompt}{Key Prompt Qualities}
    \begin{itemize}[wide]
    \item \textbf{Specificity:} Does the prompt ask for a specific, well-defined output without leaving  any ambiguity?
    \item \textbf{Domain Knowledge:} Does the prompt test the AI's knowledge and understanding in a specific domain or set of domains?
    \item \textbf{Complexity:} Does the prompt have multiple components, variables, or levels of depth and nuance? 
    \item \textbf{Problem-Solving:} Does the prompt require active problem-solving: analyzing and clearly defining the problem and systematically devising and implementing a solution? 
    \item \textbf{Creativity:} Does the prompt require a creative approach or solution? 
    \item \textbf{Technical Accuracy:} Does the prompt require an answer with a high degree of technical accuracy, correctness and precision? 
    \item \textbf{Real-world Application:} Does the prompt relate to real-world applications? 
\end{itemize}
\end{prompt}

\subsection{\dataset{}}
\label{sec:dataset_method}

We utilize the \method{} pipeline to curate 500 challenging benchmark prompts for \dataset{}. Our process begins with an initial pool of 200,000 prompts sourced from Chatbot Arena. We filter out duplicates, multi-turn conversations, and non-English content. Next, we apply hierarchical topic modeling, clustering the prompts into 4,000 distinct topics spanning a diverse range of domains

Then we use GPT-4-Turbo~\citep{openai2023gpt4turbo} as a judge to assign a “quality score” to each prompt and remove any prompts. Prompts with score less than 6 and topic clusters with mean score less than 5 are discarded, ensuring only the highest quality prompts are retained. The resulting dataset contains over 500 high quality clusters. To construct a 500-prompt benchmark, we sample 2 prompts each from 250 randomly selected clusters. We also ensure the final dataset is free from personally identifiable information or offensive content.

To validate qualities assigned by GPT-4-Turbo, we construct “ground truth” labels for 200 sampled queries by collecting majority votes from GPT-4o~\citep{openai2024gpt4o}, Claude-3-Opus, and Gemini-1.5-Pro~\citep{reid2024gemini}. GPT-4-Turbo achieves 85.6\% agreement with these labels, demonstrating its reliability as an annotator.

We also applied \method{} on 150,000 queries from WildChat-1M~\citep{zhao2024wildchat}, which consists of diverse and real-world conversations between users and ChatGPT. \method{} identified 185 high quality clusters with 4,500+ prompts. We then randomly sample 2 prompts from each of the highest-quality 125 clusters to create a new benchmark, \wildset{}, which we show to have similar improvement in benchmark quality in section~\ref{sec:6.4}.

\subsection{Pipeline Cost and Statistic Analysis}
\label{sec:pipeline_cost_analysis}
The estimated cost for applying \method{} on 200,000 Chatbot Arena queries using GPT-4-Turbo as annotator is approximately \$500~\footnote{250 tokens per prompt on average x 200,000 user queries x \$10 per 1 million tokens (OpenAI pricing for GPT-4-1106-Preview).}. This cost can be significantly reduced if employing Llama-3-70B-Instruct~\citep{dubey2024llama} as annotator instead, which only cost around \$45~\footnote{250 tokens per prompt on average x 200,000 user queries x \$0.9 per 1 million tokens (TogetherAI pricing, date: 2024-10-01).}. We experimented with Llama-3-70B-Instruct as an alternative annotator and observed similar improvement in downstream benchmark quality. Results are discussed in section~\ref{sec:6.4}.

\begin{figure}[t]
    \centering
    \begin{minipage}[b]{0.48\textwidth}
        \centering
        \includegraphics[width=\textwidth]{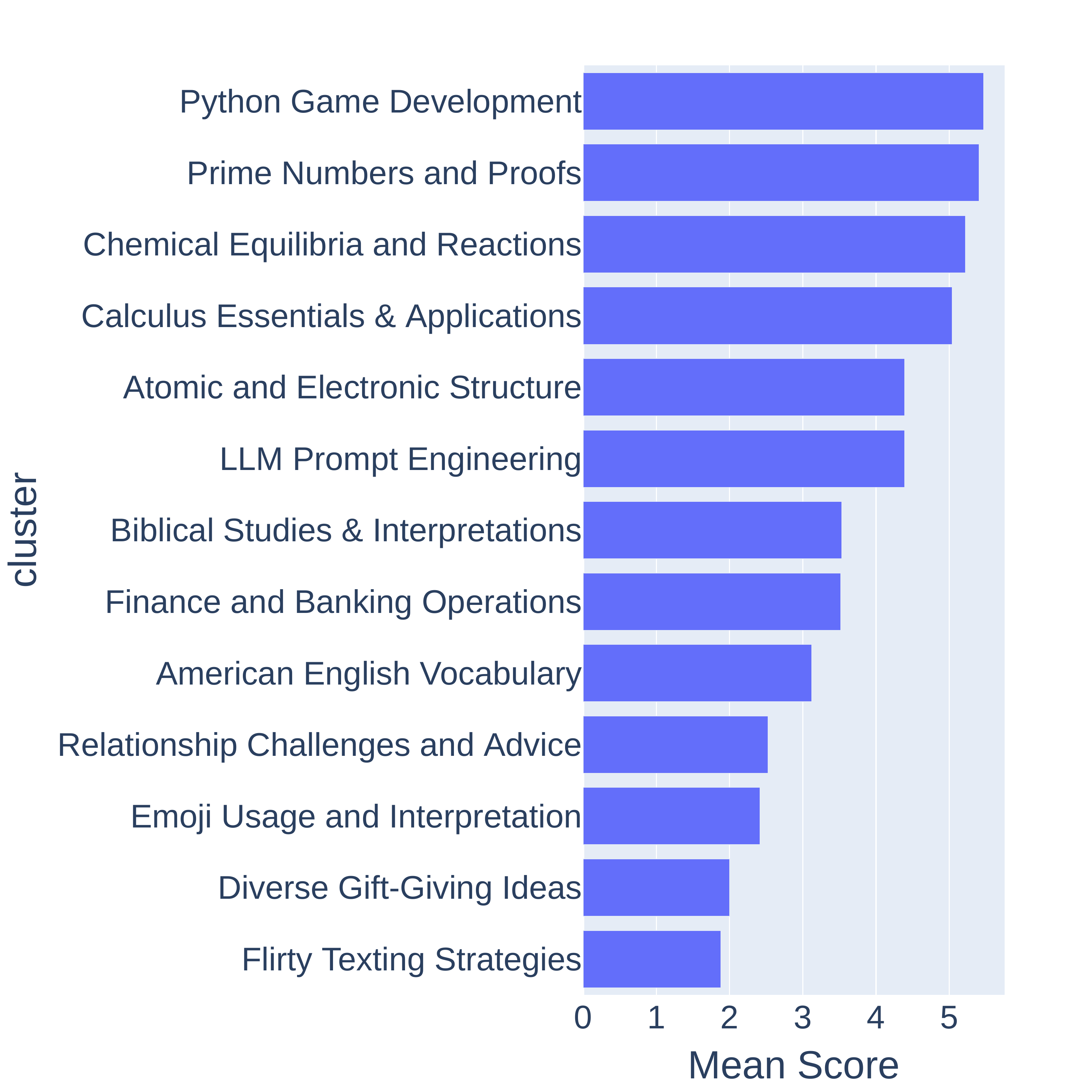}
        \caption{Mean score of various topic clusters in descending order. Higher-scoring clusters correlate to challenging topics. A more complete topic cluster plot is in \autoref{fig:cluster_distribution_full}.}
        \label{fig:cluster_distribution}
    \end{minipage}
    \hfill
    \begin{minipage}[b]{0.48\textwidth}
        \centering
        \includegraphics[width=\textwidth]{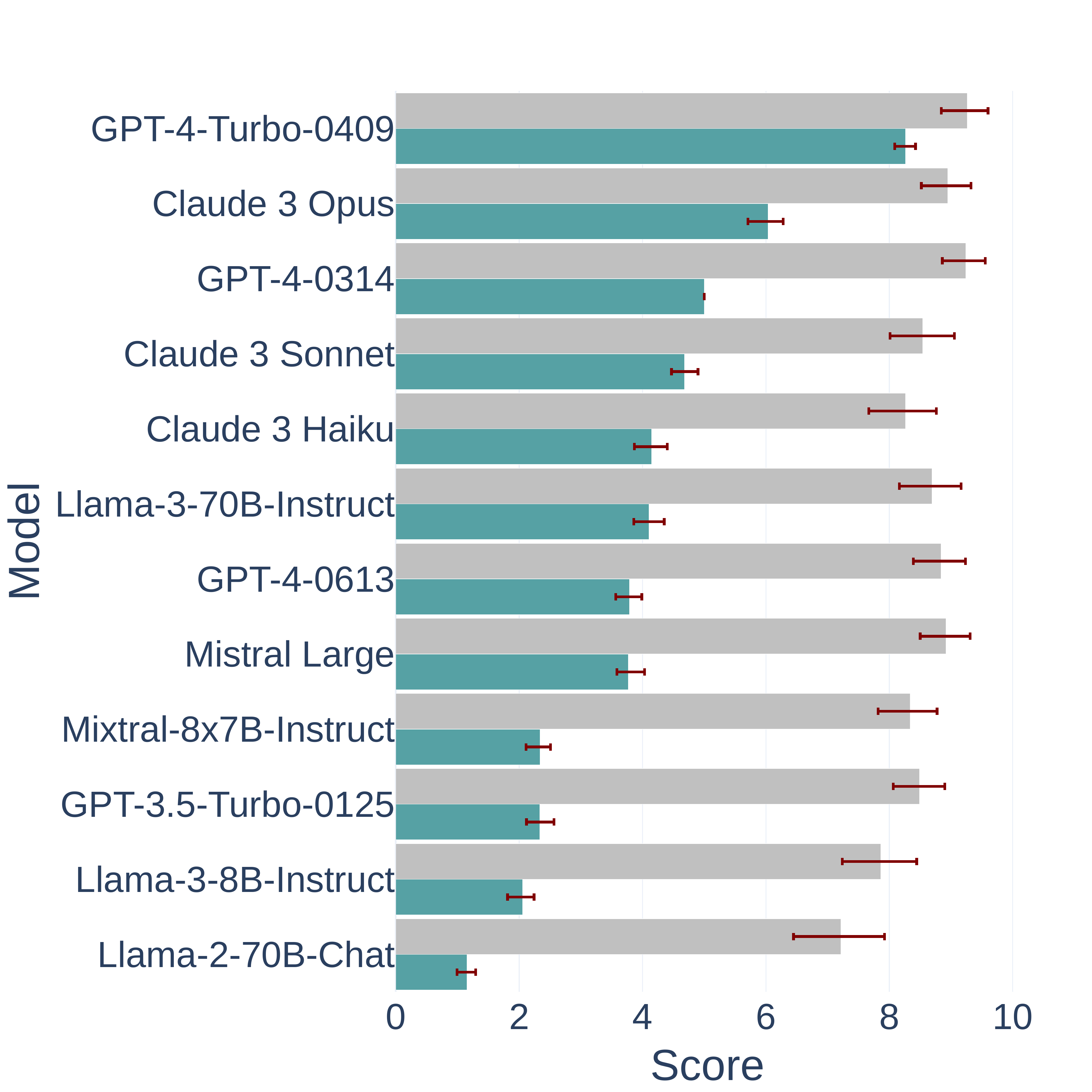}
        \caption{Comparison between \dataset{} (Green) and MT-Bench (Grey). The former offers significantly better separability between models and tighter confidence intervals.}
        \label{fig:mt_bench_arenahard}
    \end{minipage}
\end{figure}

Figure~\ref{fig:cluster_distribution} illustrates examples of topic clusters across a spectrum of mean scores. Clusters with higher scores correspond to complex topics such as game development or mathematical proofs, while lower-scoring clusters typically involve simpler or ambiguous questions (e.g., "Flirty Texting Strategies"). We provide further examples of prompts and their respective topic clusters in Appendix~\ref{sec:examples}.

\begin{figure}[t]
    \centering
    \includegraphics[width=1.0\textwidth]{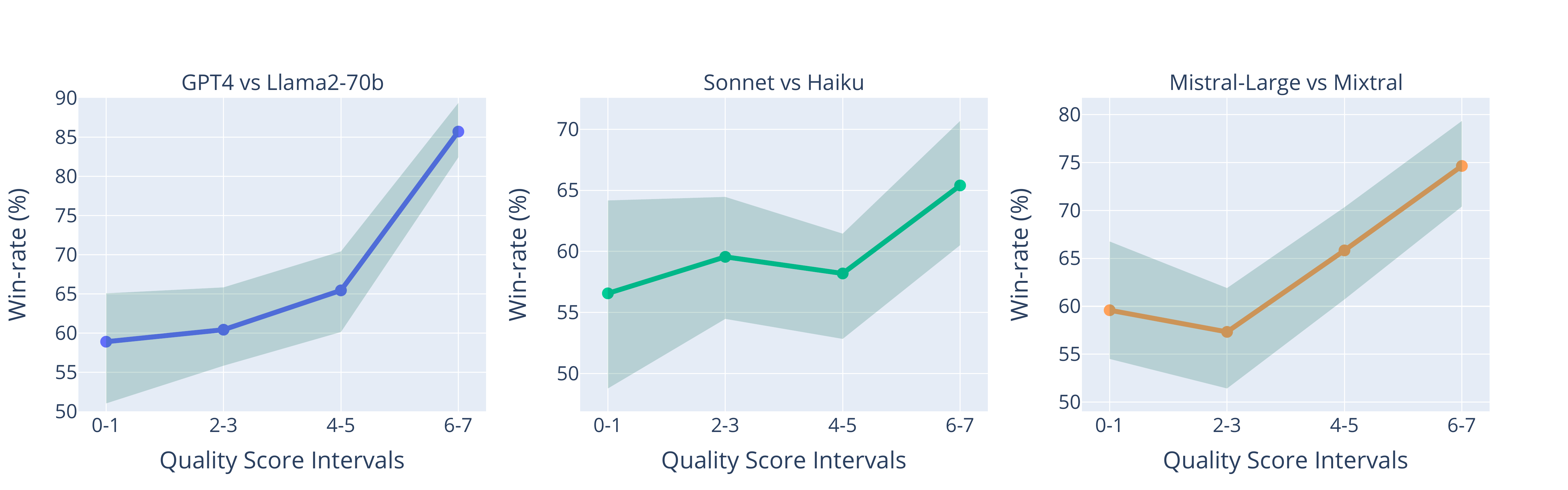}
    \caption{Win-rate of three model pairs
    (GPT-4-0613 vs Llama-2-70b-chat, Claude-3-Sonnet-20240229 vs Claude-3-Haiku-20240307, and Mistral-Large vs Mixtral-8x7b-Instruct-v0.1)
    over ``quality score''. We randomly sample 50 queries for each quality score 0-7 and bootstrap a win-rate and confidence interval between model pairs on each score interval of 2. We observe a similar trend of win-rate between model pairs becomes increasingly separable as the quality score increases.}
    \label{fig:score_separability_full}
    \vspace{-0.5em}
\end{figure}

To see whether ``quality score'' assigned during \method{}'s pipeline correlates with separability and agreement, we sample 50 prompts per score and compare the responses from GPT-4 and Llama-2-70b-Chat~\citep{touvron2023llama2}, with GPT-4-Turbo as judge. In Figure~\ref{fig:score_separability_full} (Left), we observe a strong correlation between high potential score and the win-rate of GPT-4-Turbo over Llama-2-70b-Chat. Similar trends are across other model pairs, including Claude Sonnet vs Haiku and Mistral-Large~\citep{mistral2024large} vs Mixtral~\citep{jiang2024mixtral}.

\section{Evaluation with LLM-as-a-Judge}
\label{sec:evaluation}
Evaluating models on challenging queries such as \dataset{} requires expert-level judgment due to the depth of domain knowledge and problem-solving skills involved. Expert evaluation, while ideal, is both costly and time-consuming. To address this, we leverage the LLM-as-a-Judge framework~\citep{llmjudge,dubois2023alpacafarm} as a scalable alternative to approximate human preferences.

We evaluate a model on a given prompt using a pairwise comparison against a strong baseline model (e.g., GPT-4-0314). A judge model (e.g., GPT-4-Turbo or Gemini-1.5-Pro) then scores each output by rating its preference between the pair on a 5-point Likert scale~\citep{likert1932technique} (1 indicates strong preference for model A, 5 indicates strong preference for model B). This scoring method penalizes models more heavily for large losses, effectively distinguishing performance across models. To ensure consistency, we utilize chain-of-thought~\citep{wei2023chainofthought} prompting, guiding the LLM judge to generate its own solution before issuing a judgment. Detailed prompt templates are provided in Section~\ref{sec:prompts}. To avoid potential position bias, we adopt a two-game setup – per query we swap the models on the first and second position. We also study and propose solutions to mitigate potential stylistic biases, such as answer length, and self-bias in LLM-based evaluation in section~\ref{sec:results}. 

This results in 1000 judgments per model evaluation. Following Chatbot Arena, we adopt the \citet{bradley1952rank} model to produce model’s the final model scores. We aggregate all pairwise comparisons to the baseline model for all models and bootstrapping the comparisons to retrieve a bootstrapped confidence interval of all models' win-rate against the baseline, producing a ordered ranking of all models by their win-rates. 

\section{Experimental Results}
\label{sec:results}

\begin{table}[t]
\centering
\resizebox{0.85\columnwidth}{!}{
\begin{tabular}{lrrrr}
\toprule
          & \dataset{}     & MT Bench          & AlpacaEval 2.0 LC      &  Chatbot Arena\\
\midrule
Confidence Agreement       & \textbf{90.9\%}    & 26.6\%    & 82.5\%    & N/A\\
Separability               & \textbf{87.4\%}    & 22.6\%    & 83.2\%    & 85.8\%\\
Spearman Correlation       & \textbf{93.2\%}    & 89.9\%    & 91.9\%    & N/A\\
Kendall Tau Correlation    & \textbf{80.0\%}    & 64.2\%    & 77.9\%    & N/A\\
Brier Score                & \textbf{0.069}     & 0.09      & 0.11      & N/A\\
\midrule
Real-world                 & Yes                & Mixed     & Mixed     & Yes \\
Freshness                  & Frequent Updates   & Static    & Static    & Live\\
Eval cost per model        & \$20               & \$10      & \$10      & Very High  \\
Prompts per model          & 500                & 160       & 800       & 10,000+ \\
\bottomrule
\end{tabular}}
\caption[]{We use a set of top-20 models\footnotemark on Chatbot Arena (2024/04/13) that are also present on the AlpacaEval leaderboard to calculate separability and agreement per benchmark. We consider the human preference ranking by Chatbot Arena (English only) as the reference to calculate agreement.}
\vspace{-0.5em}
\label{tab:benchmark_comparison}
\end{table}
\footnotetext{gpt-4-turbo-2024-04-09, claude-3-opus-20240229, claude-3-sonnet-20240229, gpt-4-0314~\citep{openai2023gpt4}, gpt-4-0613, mistral-large-2402, qwen1.5-72b-chat~\citep{qwen1.5}, mistral-medium, claude-2.0, gpt-3.5-turbo-0613, claude-2.1, gemini-pro~\citep{team2023gemini}, mixtral-8x7b-instruct-v0.1~\citep{jiang2024mixtralexperts}, gpt-3.5-turbo-0314, yi-34b-chat~\citep{ai2024yi}, tulu-2-dpo-70b~\citep{ivison2023camels}, dbrx-instruct-preview~\citep{dbrx2024instruct}, vicuna-33b~\citep{vicuna2023}, starling-lm-7b-alpha~\citep{starling2023}, llama-2-70b-chat~\citep{touvron2023llama2}}
\begin{table}[t]
\centering
\begin{tabular}{lrr}
\toprule
          & \wildset{} & Wild-Random-250\\
\midrule

Confidence Agreement & 88.6\% & 36.4\% \\
Separability & 86.7\% & 75.6\% \\
Spearman Correlation & 91.5\% & 45.5\% \\
\bottomrule
\end{tabular}
\vspace{0.1em}
\caption{Comparing \wildset{} and a baseline of 250 prompts randomly selected from the WildChat dataset, using GPT-4-Turbo as the judge. \wildset{} has significantly higher separability and agreement to human preference ranking. The experiment demonstrates BenchBuilder's robustness as a general data curation pipeline across different datasets.}
\label{tab:comparison_wild_hard}
\vspace{-1.0em}
\end{table}

\begin{table}[t]
\centering
\resizebox{1\columnwidth}{!}{
\begin{tabular}{lrrrr}
\toprule
          & \dataset{} (Style Control) & \dataset{} & AlpacaEval 2.0 LC & MT-Bench \\
\midrule
Confidence Agreement  & \textbf{98.6\%}  & 94.4\%           & 83.8\% & 30.3\% \\
Separability            & 86.8\%           & \textbf{87.4\%}  & 83.2\% & 22.6\% \\
Spearman Correlation    & \textbf{98.6\%}  & 94.9\%           & 88.1\% & 90.7\% \\
Kendall Tau Correlation & \textbf{93.7\%}  & 85.3\%           & 70.5\% & 77.9\% \\
\bottomrule
\end{tabular}}
\vspace{0.1em}
\caption{We apply style control to Chatbot Arena battles (English Hard Prompts) and use its model ranking as reference to calculate alignment. When stylistic confounders like response length are controlled, \dataset{} achieves high alignment to human preferences.  }
\vspace{-0.5em}
\label{tab:benchmark_comparison_style_control}
\end{table}

\begin{table}[t]
\centering
\resizebox{1\columnwidth}{!}{
\begin{tabular}{lrrrrr}
\toprule
Model          & GPT4-T & Claude3-Opus & Gemini1.5-Pro & Llama3-70B & Ensemble-as-Judges\\
\midrule
Confiderence Agreement     & 90.9\% & 66.7\% & 84.8\% & 65.6\% & \textbf{91.5\%}\\
Separability   & 87.4\% & 83.68\% & 82.11\% & 81.6\% & \textbf{89.5\%}\\
Spearman Correlation   & 93.2\% & 77.0\% & 95.2\% & 70.5\% & \textbf{96.5\%}\\
Brier Score   & 0.069 & 0.170 & \textbf{0.064} & 0.196 & 0.065\\
\bottomrule
\end{tabular}}
\vspace{0.1em}
\caption{Statistics of \dataset{} with four LLM different judges: GPT4-T (gpt-4-1106-preview), Claude-3-Opus, Gemini1.5-Pro (gemini-1.5-pro-0514), Llama3-70B (llama-3-70b-instruct). We compare rankings produced by these judges against Chatbot Arena (English) ranking (as of 2024/04/13). We observe GPT-4T and Gemini1.5-Pro have higher agreement than Claude-3-Opus and Llama-3-70B. Furthermore, the ensemble of GPT4-T and Gemini1.5-Pro shows even higher agreement.}
\vspace{-0.5em}
\label{tab:benchmark_judge_comparison}
\end{table}

\begin{table}[ht]
\centering
\begin{minipage}{.5\textwidth}
\centering
\caption*{No Modification}
\begin{tabular}{lr}
\toprule
Model & Score \\
\midrule
Llama-3.1-70B-Instruct-detail & 53.5 \\
Llama-3.1-70B-Instruct-md & 44.9 \\
Llama-3.1-70B-Instruct & 44.5 \\
Llama-3.1-70B-Instruct-chatty & 44.3 \\
Llama-3.1-70B-Instruct-no-md & 37.5 \\
\bottomrule
\end{tabular}
\end{minipage}%
\begin{minipage}{.5\textwidth}
\centering
\caption*{Style Control}
\begin{tabular}{lr}
\toprule
Model & Score \\
\midrule
Llama-3.1-70B-Instruct & 41.7 \\
Llama-3.1-70B-Instruct-no-md & 39.9 \\
Llama-3.1-70B-Instruct-detail & 39.8 \\
Llama-3.1-70B-Instruct-chatty & 39.5 \\
Llama-3.1-70B-Instruct-md & 34.9 \\
\bottomrule
\end{tabular}
\end{minipage}
\caption{Comparison Between \dataset{} with no modification versus applying style control. Left: \dataset{} with no modification to GPT-4-Turbo judge. Right: style controlled GPT-4-Turbo judge. Asking Llama-3.1-70B-Instruct~\citep{dubey2024llama} to response with more detail shows significant performance gain when no style control is applied. However, it is no longer favored with style control. Full table with additional models and system instructions can be found in Appendix Table~\ref{tab:comparison_style_control_causal_full}.}
\label{tab:comparison_style_control_causal}
\end{table}

\subsection{Setup and Baselines}

To compare \dataset{}'s separability and alignment with humans against other widely used benchmarks, MT-Bench~\citep{llmjudge} and AlpacaEval 2.0 Length Controlled~\citep{dubois2024length}, we obtain 95\% confidence intervals of model performances via applying 100 rounds of bootstrapping on judgment results for each benchmark. For AlpacaEval, we use pre-existing results from their repository. We obtain MT-Bench judgment with no modification to their recommended evaluation setup. For \dataset{}, we employ the system proposed in \autoref{sec:evaluation} by choosing gpt-4-0314 as baseline model for pairwise comparison.

To ensure fair comparison, we use a set of top-20 models\textsuperscript{3} on Chatbot Arena \citep{chiang2024chatbot} (2024/04/13) that are also presented on AlpacaEval leaderboard (2024/04/13) as ground truth for human preferences on the model ranking orders.

\subsection{Comparing Separability and Alignment Across Benchmarks}
In Table~\ref{tab:benchmark_comparison}, \dataset{} shows the highest separability (87.4\%) against widely adopted LLM benchmarks and offers highest agreement (90.8\%) to Chatbot Arena at a \$20 cost. In Figure~\ref{fig:mt_bench_arenahard}, we show \dataset{} offers significantly stronger separability against MT-Bench with tighter confidence intervals. With only 500 prompts, \dataset{} achieve impressive alignment to (and even higher separability than) Chatbot Arena Rankings, which constitutes over 1 million real-world human preferences.

Notably, we observe a significant gap between MT-bench’s Spearman Correlation (89.9\%) and confidence agreement (22.6\%) to Chatbot Arena, an example where Spearman Correlation fails to account for variance of the rankings, and hence cannot adequately measure important ranking granularity of top LLMs. We present a visual comparison between \dataset{} and MT-Bench in \autoref{fig:mt_bench_arenahard}, highlighting \dataset{}'s improved separability.

\subsection{Comparing to a similar distribution of human preference} 
\label{in_distribution_results}
We evaluate \dataset{} with Chatbot Arena's English Hard Prompt leaderboard as ground truth. Since this version of Chatbot Arena leaderboard is based on votes from a more challenging subset of the overall Chatbot Arena battles, we believe it is a more in-distribution comparison for \dataset{}, which also consist of challenging user queries. We observe \dataset{} achieves an overall higher alignment (98.6\% Confidence Agreement and 96.7\% Spearman Correlation) to human preferences. Results are presented in Appendix Table~\ref{tab:comparison_hard_category}.

\subsection{Robustness and Generalizability}
\label{sec:6.4}
To evaluate the robustness and generalizability of the \method{} pipeline, we applied it on 150,000 WildChat~\citep{zhao2024wildchat} dataset and identified 185 high quality clusters with 4,500+ prompts. We then randomly sample 2 prompts from each of the highest-quality 125 clusters to create a new benchmark, \wildset{}. We compare \wildset{} and a baseline of 250 prompts randomly selected from the WildChat dataset in table~\ref{tab:comparison_wild_hard}. Results indicates \wildset{} has significantly higher separability and agreement to human preference ranking than a random baseline, demonstrating \method{}'s robustness as a general data curation pipeline for various crowdsourced datasets.

Additionally, we compared \dataset{} against two separate sets of 500 randomly selected prompts from the Chatbot Arena dataset, prior to applying the pipeline extraction. We observe \dataset{} significantly outperforms both random baselines. Results are shown in Appendix Table~\ref{tab:random_baseline_comparison}.

To verify whether \method{} is not limited to GPT-4-Turbo as annotator for prompt qualities, we employed Llama-3-70B-Instruct as an alternative annotator for prompt curation. We observe the benchmark produced by Llama-3-70b-instruct as the prompt annotator has similar improvement in quality as \dataset{} from random baselines. Results are shown in Appendix Table~\ref{tab:comparison_llama_annotator}.

\subsection{Mitigating Stylistic Biases in LLM-Based Evaluation}
LLM-as-a-Judge based evaluation is known to suffer from various biases, such as favoring longer responses~\citep{llmjudge, dubois2024length}. AlpacaEval 2.0 Length Control~\citep{dubois2024length} proposes an regression based approach to control length bias in LLM-based evaluation. Chatbot Arena also released a style controlled leaderboard~\citep{stylearena2024}, which attempts to decouple substance from stylistic preferences, including answer length and markdown usage. Following their approaches, we modify how \dataset{} computes the model scores by accounting for the stylistic differences between two answers as additional features to the existing Bradley-Terry model.

We propose controlling for a similar set of stylistic elements used to control human preference on Chatbot Arena for LLM-based evaluation: \textbf{answer token length}, density of \textbf{markdown headers}, \textbf{markdown bold elements}, and \textbf{markdown lists}. Technical details on how to extend the Bradley-Terry model for controlling any given style can be found in Appendix~\ref{appendix:A.2}. 

We apply style control to Chatbot Arena battles and compare the resulting model preference ranking to style controlled \dataset{}, aiming to answer the question: \textit{How well aligned is \dataset{} to human preference when both human preference and LLM judgment are decoupled from stylistic differences?} In Table~\ref{tab:benchmark_comparison_style_control}, we show that style controlled \dataset{} achieves \textbf{98.6\% agreement and correlation} to style controlled human preference ranking, suggesting \dataset{} assessment of model strength separated from style is still highly aligned to humans.

Additionally, we conducted an experiment trying to increase model score on \dataset{} by instructing GPT-3.5-Turbo, Llama-3.1-70b-instruct, and Gemini-1.5-Flash to increase the verbosity and usage of markdown elements in their response and present our results in Table~\ref{tab:comparison_style_control_causal}. While increasing ``detailedness'' does increase model performances on \dataset{} when no modifications is applied to GPT-4-Turbo as judge, applying style control is effective at neutralizing this advantage. Our results shows that style controlled model scores cannot be gamed via manipulating response length or markdown usage on \dataset{}. We also observe a reduction in correlation between model score and answer length on \dataset{}. Full results can be found in Appendix Table~\ref{tab:comparison_style_control_verbosity_correlation}. 

\subsection{Mitigating Self-Biases in LLM-Based Evaluation}
LLM-as-a-Judge evaluations are also known to exhibit self-bias. While such biases should manifest as lower alignment with human preferences in our proposed metrics, we conduct a focused analysis to further understand and address this issue. Since \dataset{} uses GPT-4-Turbo as the default judge, we evaluate whether it favors OpenAI models over Anthropic models. Results in Appendix Table~\ref{tab:comparison_self_bias} indicate that GPT models receive slightly higher average rankings than human preference, while Claude models rank lower.

To reduce this bias, we propose Ensemble-as-Judges, which aggregates judgments from multiple models. The ensemble judges (GPT-4-Turbo and Gemini-1.5-Pro) achieves overall higher separability and alignment with human rankings, as shown in Table~\ref{tab:benchmark_judge_comparison}. Additionally, we also observe that combining GPT-4-Turbo and Gemini-1.5-Pro reduces self-biases. Results can be found in Appendix Table~\ref{tab:comparison_self_bias}. We believe further research into ensemble methods can refine these results and leave this for future exploration.



\section{Limitations}
\label{sec:limitations}
While our data sources are drawn from diverse distributions, biases may still exist in our pipeline. For instance, the seven defined qualities may not fully capture the range of possible attributes, potentially skewing towards prompts in technical domains. Furthermore, \dataset{} currently lacks evaluation for multi-turn and non-English interactions due to the limited availability of multi-turn data in crowdsourced datasets and the primary language proficiency of the authors.

To address these limitations, future work will focus on expanding \method{} to incorporate multi-turn and multilingual data curation. We also aim to refine our prompt quality definitions, creating a more systematic approach for generating benchmarks that reflect a broader, more inclusive range of scenarios while maintaining high separability and alignment with human judgment. We also plan to explore more advanced version of Ensemble-as-Judges to further enhance our LLM-based evaluation approach.

\section{Conclusions}

We introduced \method{}, a data curation pipeline that transforms crowdsourced data into high-quality benchmarks by seven key qualities. This pipeline enables building challenging and evolving benchmarks which is crucial for evaluating today's advanced language models. 
Our evaluation metrics, including separability and agreement with confidence, provide a comprehensive assessment of benchmarks.
We show the resulting benchmark, \dataset{}, significantly improves separability and alignment with human preferences over existing benchmarks, achieving 98.6\% agreement with Chatbot Arena rankings at only \$20 per evaluation. 
We expect \dataset{} to be useful for LLM developers to evaluate their models with confidence and \method{} to be a valuable tool for developers seeking to extract high-quality benchmark from vast amounts of data with minimal human effort.




\newpage
\bibliography{iclr2025_conference}

\begin{thebibliography}{69}
\providecommand{\natexlab}[1]{#1}
\providecommand{\url}[1]{\texttt{#1}}
\expandafter\ifx\csname urlstyle\endcsname\relax
  \providecommand{\doi}[1]{doi: #1}\else
  \providecommand{\doi}{doi: \begingroup \urlstyle{rm}\Url}\fi

\bibitem[Abdin et~al.(2024)Abdin, Jacobs, Awan, Aneja, Awadallah, Awadalla, Bach, Bahree, Bakhtiari, Behl, et~al.]{abdin2024phi}
Marah Abdin, Sam~Ade Jacobs, Ammar~Ahmad Awan, Jyoti Aneja, Ahmed Awadallah, Hany Awadalla, Nguyen Bach, Amit Bahree, Arash Bakhtiari, Harkirat Behl, et~al.
\newblock Phi-3 technical report: A highly capable language model locally on your phone.
\newblock \emph{arXiv preprint arXiv:2404.14219}, 2024.

\bibitem[AI et~al.(2024)AI, :, Young, Chen, Li, Huang, Zhang, Zhang, Li, Zhu, Chen, Chang, Yu, Liu, Liu, Yue, Yang, Yang, Yu, Xie, Huang, Hu, Ren, Niu, Nie, Xu, Liu, Wang, Cai, Gu, Liu, and Dai]{ai2024yi}
01. AI, :, Alex Young, Bei Chen, Chao Li, Chengen Huang, Ge~Zhang, Guanwei Zhang, Heng Li, Jiangcheng Zhu, Jianqun Chen, Jing Chang, Kaidong Yu, Peng Liu, Qiang Liu, Shawn Yue, Senbin Yang, Shiming Yang, Tao Yu, Wen Xie, Wenhao Huang, Xiaohui Hu, Xiaoyi Ren, Xinyao Niu, Pengcheng Nie, Yuchi Xu, Yudong Liu, Yue Wang, Yuxuan Cai, Zhenyu Gu, Zhiyuan Liu, and Zonghong Dai.
\newblock Yi: Open foundation models by 01.ai, 2024.

\bibitem[Bos \& Markert(2005)Bos and Markert]{bos-markert-2005-recognising}
Johan Bos and Katja Markert.
\newblock Recognising textual entailment with logical inference.
\newblock In Raymond Mooney, Chris Brew, Lee-Feng Chien, and Katrin Kirchhoff (eds.), \emph{Proceedings of Human Language Technology Conference and Conference on Empirical Methods in Natural Language Processing}, pp.\  628--635, Vancouver, British Columbia, Canada, October 2005. Association for Computational Linguistics.
\newblock URL \url{https://aclanthology.org/H05-1079}.

\bibitem[Bowman et~al.(2015)Bowman, Angeli, Potts, and Manning]{bowman-etal-2015-large}
Samuel~R. Bowman, Gabor Angeli, Christopher Potts, and Christopher~D. Manning.
\newblock A large annotated corpus for learning natural language inference.
\newblock In Llu{\'\i}s M{\`a}rquez, Chris Callison-Burch, and Jian Su (eds.), \emph{Proceedings of the 2015 Conference on Empirical Methods in Natural Language Processing}, pp.\  632--642, Lisbon, Portugal, September 2015. Association for Computational Linguistics.
\newblock \doi{10.18653/v1/D15-1075}.
\newblock URL \url{https://aclanthology.org/D15-1075}.

\bibitem[Bradley \& Terry(1952)Bradley and Terry]{bradley1952rank}
Ralph~Allan Bradley and Milton~E Terry.
\newblock Rank analysis of incomplete block designs: I. the method of paired comparisons.
\newblock \emph{Biometrika}, 39\penalty0 (3/4):\penalty0 324--345, 1952.

\bibitem[Brier(1950)]{brier1950verification}
Glenn~W Brier.
\newblock Verification of forecasts expressed in terms of probability.
\newblock \emph{Monthly weather review}, 78\penalty0 (1):\penalty0 1--3, 1950.

\bibitem[Cai et~al.(2024)Cai, Cao, Chen, Chen, Chen, Chen, Chen, Chen, Chen, Chu, et~al.]{cai2024internlm2}
Zheng Cai, Maosong Cao, Haojiong Chen, Kai Chen, Keyu Chen, Xin Chen, Xun Chen, Zehui Chen, Zhi Chen, Pei Chu, et~al.
\newblock Internlm2 technical report.
\newblock \emph{arXiv preprint arXiv:2403.17297}, 2024.

\bibitem[Carlini et~al.(2021)Carlini, Tramer, Wallace, Jagielski, Herbert-Voss, Lee, Roberts, Brown, Song, Erlingsson, et~al.]{carlini2021extracting}
Nicholas Carlini, Florian Tramer, Eric Wallace, Matthew Jagielski, Ariel Herbert-Voss, Katherine Lee, Adam Roberts, Tom Brown, Dawn Song, Ulfar Erlingsson, et~al.
\newblock Extracting training data from large language models.
\newblock In \emph{30th USENIX Security Symposium (USENIX Security 21)}, pp.\  2633--2650, 2021.

\bibitem[Chen et~al.(2021)Chen, Tworek, Jun, Yuan, Pinto, Kaplan, Edwards, Burda, Joseph, Brockman, et~al.]{chen2021evaluating}
Mark Chen, Jerry Tworek, Heewoo Jun, Qiming Yuan, Henrique Ponde de~Oliveira Pinto, Jared Kaplan, Harri Edwards, Yuri Burda, Nicholas Joseph, Greg Brockman, et~al.
\newblock Evaluating large language models trained on code.
\newblock \emph{arXiv preprint arXiv:2107.03374}, 2021.

\bibitem[Chiang et~al.(2023)Chiang, Li, Lin, Sheng, Wu, Zhang, Zheng, Zhuang, Zhuang, Gonzalez, Stoica, and Xing]{vicuna2023}
Wei-Lin Chiang, Zhuohan Li, Zi~Lin, Ying Sheng, Zhanghao Wu, Hao Zhang, Lianmin Zheng, Siyuan Zhuang, Yonghao Zhuang, Joseph~E. Gonzalez, Ion Stoica, and Eric~P. Xing.
\newblock Vicuna: An open-source chatbot impressing gpt-4 with 90\%* chatgpt quality, March 2023.
\newblock URL \url{https://lmsys.org/blog/2023-03-30-vicuna/}.

\bibitem[Chiang et~al.(2024)Chiang, Zheng, Sheng, Angelopoulos, Li, Li, Zhang, Zhu, Jordan, Gonzalez, and Stoica]{chiang2024chatbot}
Wei-Lin Chiang, Lianmin Zheng, Ying Sheng, Anastasios~Nikolas Angelopoulos, Tianle Li, Dacheng Li, Hao Zhang, Banghua Zhu, Michael Jordan, Joseph~E. Gonzalez, and Ion Stoica.
\newblock Chatbot arena: An open platform for evaluating llms by human preference, 2024.

\bibitem[Cobbe et~al.(2021)Cobbe, Kosaraju, Bavarian, Chen, Jun, Kaiser, Plappert, Tworek, Hilton, Nakano, et~al.]{cobbe2021gsm}
Karl Cobbe, Vineet Kosaraju, Mohammad Bavarian, Mark Chen, Heewoo Jun, Lukasz Kaiser, Matthias Plappert, Jerry Tworek, Jacob Hilton, Reiichiro Nakano, et~al.
\newblock Training verifiers to solve math word problems.
\newblock \emph{arXiv preprint arXiv:2110.14168}, 2021.

\bibitem[DeepSeek-AI et~al.(2024)DeepSeek-AI, Zhu, Guo, Shao, Yang, Wang, Xu, Wu, Li, Gao, Ma, Zeng, Bi, Gu, Xu, Dai, Dong, Zhang, Piao, Gou, Xie, Hao, Wang, Song, Chen, Xie, Guan, You, Liu, Du, Gao, Lu, Chen, Wang, Deng, Li, Zhao, Ruan, Luo, and Liang]{deepseekai2024deepseekcoderv2breakingbarrierclosedsource}
DeepSeek-AI, Qihao Zhu, Daya Guo, Zhihong Shao, Dejian Yang, Peiyi Wang, Runxin Xu, Y.~Wu, Yukun Li, Huazuo Gao, Shirong Ma, Wangding Zeng, Xiao Bi, Zihui Gu, Hanwei Xu, Damai Dai, Kai Dong, Liyue Zhang, Yishi Piao, Zhibin Gou, Zhenda Xie, Zhewen Hao, Bingxuan Wang, Junxiao Song, Deli Chen, Xin Xie, Kang Guan, Yuxiang You, Aixin Liu, Qiushi Du, Wenjun Gao, Xuan Lu, Qinyu Chen, Yaohui Wang, Chengqi Deng, Jiashi Li, Chenggang Zhao, Chong Ruan, Fuli Luo, and Wenfeng Liang.
\newblock Deepseek-coder-v2: Breaking the barrier of closed-source models in code intelligence, 2024.
\newblock URL \url{https://arxiv.org/abs/2406.11931}.

\bibitem[Dolan \& Brockett(2005)Dolan and Brockett]{dolan2005automatically}
Bill Dolan and Chris Brockett.
\newblock Automatically constructing a corpus of sentential paraphrases.
\newblock In \emph{Third international workshop on paraphrasing (IWP2005)}, 2005.

\bibitem[Dua et~al.(2019)Dua, Wang, Dasigi, Stanovsky, Singh, and Gardner]{dua2019drop}
Dheeru Dua, Yizhong Wang, Pradeep Dasigi, Gabriel Stanovsky, Sameer Singh, and Matt Gardner.
\newblock Drop: A reading comprehension benchmark requiring discrete reasoning over paragraphs.
\newblock In \emph{Proceedings of the 2019 Conference of the North American Chapter of the Association for Computational Linguistics: Human Language Technologies, Volume 1 (Long and Short Papers)}, pp.\  2368--2378, 2019.

\bibitem[Dubey et~al.(2024)Dubey, Jauhri, Pandey, Kadian, Al-Dahle, Letman, Mathur, Schelten, Yang, Fan, et~al.]{dubey2024llama}
Abhimanyu Dubey, Abhinav Jauhri, Abhinav Pandey, Abhishek Kadian, Ahmad Al-Dahle, Aiesha Letman, Akhil Mathur, Alan Schelten, Amy Yang, Angela Fan, et~al.
\newblock The llama 3 herd of models.
\newblock \emph{arXiv preprint arXiv:2407.21783}, 2024.

\bibitem[Dubois et~al.(2023)Dubois, Li, Taori, Zhang, Gulrajani, Ba, Guestrin, Liang, and Hashimoto]{dubois2023alpacafarm}
Yann Dubois, Xuechen Li, Rohan Taori, Tianyi Zhang, Ishaan Gulrajani, Jimmy Ba, Carlos Guestrin, Percy Liang, and Tatsunori~B. Hashimoto.
\newblock Alpacafarm: A simulation framework for methods that learn from human feedback, 2023.

\bibitem[Dubois et~al.(2024)Dubois, Galambosi, Liang, and Hashimoto]{dubois2024length}
Yann Dubois, Bal{\'a}zs Galambosi, Percy Liang, and Tatsunori~B Hashimoto.
\newblock Length-controlled alpacaeval: A simple way to debias automatic evaluators.
\newblock \emph{arXiv preprint arXiv:2404.04475}, 2024.

\bibitem[Frick et~al.(2024)Frick, Jin, Li, Ganesan, Zhang, Jiao, and Zhu]{athene70b2024}
Evan Frick, Peter Jin, Tianle Li, Karthik Ganesan, Jian Zhang, Jiantao Jiao, and Banghua Zhu.
\newblock Athene-70b: Redefining the boundaries of post-training for open models, July 2024.
\newblock URL \url{https://huggingface.co/Nexusflow/Athene-70B}.

\bibitem[Gemini et~al.(2023)Gemini, Anil, Borgeaud, Wu, Alayrac, Yu, Soricut, Schalkwyk, Dai, Hauth, et~al.]{team2023gemini}
Team Gemini, Rohan Anil, Sebastian Borgeaud, Yonghui Wu, Jean-Baptiste Alayrac, Jiahui Yu, Radu Soricut, Johan Schalkwyk, Andrew~M Dai, Anja Hauth, et~al.
\newblock Gemini: a family of highly capable multimodal models.
\newblock \emph{arXiv preprint arXiv:2312.11805}, 2023.

\bibitem[GLM et~al.(2024)GLM, Zeng, Xu, Wang, Zhang, Yin, Rojas, Feng, Zhao, Lai, et~al.]{glm2024chatglm}
Team GLM, Aohan Zeng, Bin Xu, Bowen Wang, Chenhui Zhang, Da~Yin, Diego Rojas, Guanyu Feng, Hanlin Zhao, Hanyu Lai, et~al.
\newblock Chatglm: A family of large language models from glm-130b to glm-4 all tools.
\newblock \emph{arXiv preprint arXiv:2406.12793}, 2024.

\bibitem[Hendrycks et~al.(2021{\natexlab{a}})Hendrycks, Burns, Basart, Zou, Mazeika, Song, and Steinhardt]{mmlu}
Dan Hendrycks, Collin Burns, Steven Basart, Andy Zou, Mantas Mazeika, Dawn Song, and Jacob Steinhardt.
\newblock Measuring massive multitask language understanding.
\newblock \emph{Proceedings of the International Conference on Learning Representations (ICLR)}, 2021{\natexlab{a}}.

\bibitem[Hendrycks et~al.(2021{\natexlab{b}})Hendrycks, Burns, Kadavath, Arora, Basart, Tang, Song, and Steinhardt]{hendrycksmath2021}
Dan Hendrycks, Collin Burns, Saurav Kadavath, Akul Arora, Steven Basart, Eric Tang, Dawn Song, and Jacob Steinhardt.
\newblock Measuring mathematical problem solving with the math dataset.
\newblock \emph{NeurIPS}, 2021{\natexlab{b}}.

\bibitem[Ivison et~al.(2023)Ivison, Wang, Pyatkin, Lambert, Peters, Dasigi, Jang, Wadden, Smith, Beltagy, and Hajishirzi]{ivison2023camels}
Hamish Ivison, Yizhong Wang, Valentina Pyatkin, Nathan Lambert, Matthew Peters, Pradeep Dasigi, Joel Jang, David Wadden, Noah~A. Smith, Iz~Beltagy, and Hannaneh Hajishirzi.
\newblock Camels in a changing climate: Enhancing lm adaptation with tulu 2, 2023.

\bibitem[Jain et~al.(2024{\natexlab{a}})Jain, Han, Gu, Li, Yan, Zhang, Wang, Solar-Lezama, Sen, and Stoica]{jain2024livecodebench}
Naman Jain, King Han, Alex Gu, Wen-Ding Li, Fanjia Yan, Tianjun Zhang, Sida Wang, Armando Solar-Lezama, Koushik Sen, and Ion Stoica.
\newblock Livecodebench: Holistic and contamination free evaluation of large language models for code.
\newblock \emph{arXiv preprint arXiv:2403.07974}, 2024{\natexlab{a}}.

\bibitem[Jain et~al.(2024{\natexlab{b}})Jain, Shetty, Zhang, Han, Sen, and Stoica]{jain2024r2e}
Naman Jain, Manish Shetty, Tianjun Zhang, King Han, Koushik Sen, and Ion Stoica.
\newblock R2e: Turning any github repository into a programming agent environment.
\newblock In \emph{ICML}, 2024{\natexlab{b}}.

\bibitem[Jiang et~al.(2024{\natexlab{a}})Jiang, Sablayrolles, Roux, Mensch, Savary, Bamford, Chaplot, de~las Casas, Hanna, Bressand, Lengyel, Bour, Lample, Lavaud, Saulnier, Lachaux, Stock, Subramanian, Yang, Antoniak, Scao, Gervet, Lavril, Wang, Lacroix, and Sayed]{jiang2024mixtral}
Albert~Q. Jiang, Alexandre Sablayrolles, Antoine Roux, Arthur Mensch, Blanche Savary, Chris Bamford, Devendra~Singh Chaplot, Diego de~las Casas, Emma~Bou Hanna, Florian Bressand, Gianna Lengyel, Guillaume Bour, Guillaume Lample, Lélio~Renard Lavaud, Lucile Saulnier, Marie-Anne Lachaux, Pierre Stock, Sandeep Subramanian, Sophia Yang, Szymon Antoniak, Teven~Le Scao, Théophile Gervet, Thibaut Lavril, Thomas Wang, Timothée Lacroix, and William~El Sayed.
\newblock Mixtral of experts, 2024{\natexlab{a}}.

\bibitem[Jiang et~al.(2024{\natexlab{b}})Jiang, Sablayrolles, Roux, Mensch, Savary, Bamford, Chaplot, de~las Casas, Hanna, Bressand, Lengyel, Bour, Lample, Lavaud, Saulnier, Lachaux, Stock, Subramanian, Yang, Antoniak, Scao, Gervet, Lavril, Wang, Lacroix, and Sayed]{jiang2024mixtralexperts}
Albert~Q. Jiang, Alexandre Sablayrolles, Antoine Roux, Arthur Mensch, Blanche Savary, Chris Bamford, Devendra~Singh Chaplot, Diego de~las Casas, Emma~Bou Hanna, Florian Bressand, Gianna Lengyel, Guillaume Bour, Guillaume Lample, Lélio~Renard Lavaud, Lucile Saulnier, Marie-Anne Lachaux, Pierre Stock, Sandeep Subramanian, Sophia Yang, Szymon Antoniak, Teven~Le Scao, Théophile Gervet, Thibaut Lavril, Thomas Wang, Timothée Lacroix, and William~El Sayed.
\newblock Mixtral of experts, 2024{\natexlab{b}}.
\newblock URL \url{https://arxiv.org/abs/2401.04088}.

\bibitem[Jimenez et~al.(2024)Jimenez, Yang, Wettig, Yao, Pei, Press, and Narasimhan]{jimenez2024swebench}
Carlos~E. Jimenez, John Yang, Alexander Wettig, Shunyu Yao, Kexin Pei, Ofir Press, and Karthik Narasimhan.
\newblock Swe-bench: Can language models resolve real-world github issues?
\newblock In \emph{The Twelfth International Conference on Learning Representations}, 2024.
\newblock URL \url{https://openreview.net/forum?id=VTF8yNQM66}.

\bibitem[Karpinska et~al.(2021)Karpinska, Akoury, and Iyyer]{karpinska-etal-2021-perils}
Marzena Karpinska, Nader Akoury, and Mohit Iyyer.
\newblock The perils of using {M}echanical {T}urk to evaluate open-ended text generation.
\newblock In Marie-Francine Moens, Xuanjing Huang, Lucia Specia, and Scott Wen-tau Yih (eds.), \emph{Proceedings of the 2021 Conference on Empirical Methods in Natural Language Processing}, pp.\  1265--1285, Online and Punta Cana, Dominican Republic, November 2021. Association for Computational Linguistics.
\newblock \doi{10.18653/v1/2021.emnlp-main.97}.
\newblock URL \url{https://aclanthology.org/2021.emnlp-main.97}.

\bibitem[Kiela et~al.(2021)Kiela, Bartolo, Nie, Kaushik, Geiger, Wu, Vidgen, Prasad, Singh, Ringshia, Ma, Thrush, Riedel, Waseem, Stenetorp, Jia, Bansal, Potts, and Williams]{dynabench2021}
Douwe Kiela, Max Bartolo, Yixin Nie, Divyansh Kaushik, Atticus Geiger, Zhengxuan Wu, Bertie Vidgen, Grusha Prasad, Amanpreet Singh, Pratik Ringshia, Zhiyi Ma, Tristan Thrush, Sebastian Riedel, Zeerak Waseem, Pontus Stenetorp, Robin Jia, Mohit Bansal, Christopher Potts, and Adina Williams.
\newblock Dynabench: Rethinking benchmarking in nlp.
\newblock \emph{NAACL}, 2021.

\bibitem[Li et~al.(2024)Li, Angelopoulos, and Chiang]{stylearena2024}
Tianle Li, Anastasios Angelopoulos, and Wei-Lin Chiang.
\newblock Does style matter? disentangling style and substance in chatbot arena, August 2024.
\newblock URL \url{https://blog.lmarena.ai/blog/2024/style-control/}.

\bibitem[Li et~al.(2023)Li, Zhang, Dubois, Taori, Gulrajani, Guestrin, Liang, and Hashimoto]{alpaca_eval}
Xuechen Li, Tianyi Zhang, Yann Dubois, Rohan Taori, Ishaan Gulrajani, Carlos Guestrin, Percy Liang, and Tatsunori~B. Hashimoto.
\newblock Alpacaeval: An automatic evaluator of instruction-following models.
\newblock \url{https://github.com/tatsu-lab/alpaca_eval}, 2023.

\bibitem[Liang et~al.(2022)Liang, Bommasani, Lee, Tsipras, Soylu, Yasunaga, Zhang, Narayanan, Wu, Kumar, et~al.]{liang2022holistic}
Percy Liang, Rishi Bommasani, Tony Lee, Dimitris Tsipras, Dilara Soylu, Michihiro Yasunaga, Yian Zhang, Deepak Narayanan, Yuhuai Wu, Ananya Kumar, et~al.
\newblock Holistic evaluation of language models.
\newblock \emph{arXiv preprint arXiv:2211.09110}, 2022.

\bibitem[Likert(1932)]{likert1932technique}
Rensis Likert.
\newblock A technique for the measurement of attitudes.
\newblock \emph{Archives of psychology}, 1932.

\bibitem[Lin et~al.(2024)Lin, Deng, Chandu, Brahman, Ravichander, Pyatkin, Dziri, Bras, and Choi]{lin2024wildbench}
Bill~Yuchen Lin, Yuntian Deng, Khyathi Chandu, Faeze Brahman, Abhilasha Ravichander, Valentina Pyatkin, Nouha Dziri, Ronan~Le Bras, and Yejin Choi.
\newblock Wildbench: Benchmarking llms with challenging tasks from real users in the wild.
\newblock \emph{arXiv preprint arXiv:2406.04770}, 2024.

\bibitem[Liu et~al.(2023)Liu, Yu, Zhang, Xu, Lei, Lai, Gu, Ding, Men, Yang, Zhang, Deng, Zeng, Du, Zhang, Shen, Zhang, Su, Sun, Huang, Dong, and Tang]{liu2023agentbench}
Xiao Liu, Hao Yu, Hanchen Zhang, Yifan Xu, Xuanyu Lei, Hanyu Lai, Yu~Gu, Hangliang Ding, Kaiwen Men, Kejuan Yang, Shudan Zhang, Xiang Deng, Aohan Zeng, Zhengxiao Du, Chenhui Zhang, Sheng Shen, Tianjun Zhang, Yu~Su, Huan Sun, Minlie Huang, Yuxiao Dong, and Jie Tang.
\newblock Agentbench: Evaluating llms as agents, 2023.

\bibitem[Ni et~al.(2024)Ni, Xue, Yue, Deng, Shah, Jain, Neubig, and You]{ni2024mixeval}
Jinjie Ni, Fuzhao Xue, Xiang Yue, Yuntian Deng, Mahir Shah, Kabir Jain, Graham Neubig, and Yang You.
\newblock Mixeval: Deriving wisdom of the crowd from llm benchmark mixtures.
\newblock \emph{arXiv preprint arXiv:2406.06565}, 2024.

\bibitem[OpenAI(2023{\natexlab{a}})]{openai2023gpt4}
OpenAI.
\newblock Gpt-4 technical report.
\newblock \emph{arXiv preprint arXiv:2303.08774}, 2023{\natexlab{a}}.

\bibitem[OpenAI(2023{\natexlab{b}})]{openai2023gpt4turbo}
OpenAI.
\newblock New models and developer products announced at devday.
\newblock \url{https://openai.com/blog/new-models-and-developer-products-announced-at-devday}, 2023{\natexlab{b}}.
\newblock (Accessed on 06/05/2024).

\bibitem[OpenAI(2024{\natexlab{a}})]{openai2024embedding}
OpenAI.
\newblock New embedding models and api updates.
\newblock \url{https://openai.com/index/new-embedding-models-and-api-updates/}, 2024{\natexlab{a}}.
\newblock (Accessed on 06/05/2024).

\bibitem[OpenAI(2024{\natexlab{b}})]{openai2024gpt4o}
OpenAI.
\newblock Hello gpt-4o.
\newblock \url{https://openai.com/index/hello-gpt-4o/}, 2024{\natexlab{b}}.
\newblock (Accessed on 06/05/2024).

\bibitem[Pearson(1895)]{pearson}
Karl Pearson.
\newblock Note on regression and inheritance in the case of two parents.
\newblock \emph{Proceedings of the Royal Society of London}, 58:\penalty0 240--242, 1895.
\newblock ISSN 03701662.
\newblock URL \url{http://www.jstor.org/stable/115794}.

\bibitem[Rajpurkar et~al.(2016)Rajpurkar, Zhang, Lopyrev, and Liang]{rajpurkar2016squad}
Pranav Rajpurkar, Jian Zhang, Konstantin Lopyrev, and Percy Liang.
\newblock Squad: 100,000+ questions for machine comprehension of text.
\newblock \emph{EMNLP}, 2016.

\bibitem[Reid et~al.(2024)Reid, Savinov, Teplyashin, Lepikhin, Lillicrap, Alayrac, Soricut, Lazaridou, Firat, Schrittwieser, et~al.]{reid2024gemini}
Machel Reid, Nikolay Savinov, Denis Teplyashin, Dmitry Lepikhin, Timothy Lillicrap, Jean-baptiste Alayrac, Radu Soricut, Angeliki Lazaridou, Orhan Firat, Julian Schrittwieser, et~al.
\newblock Gemini 1.5: Unlocking multimodal understanding across millions of tokens of context.
\newblock \emph{arXiv preprint arXiv:2403.05530}, 2024.

\bibitem[Rein(2024)]{gpqacostblog2024}
David Rein.
\newblock Can good benchmarks contain mistakes?, 2024.
\newblock URL \url{https://wp.nyu.edu/arg/can-good-benchmarks-contain-mistakes/}.

\bibitem[Rein et~al.(2023)Rein, Hou, Stickland, Petty, Pang, Dirani, Michael, and Bowman]{rein2023gpqa}
David Rein, Betty~Li Hou, Asa~Cooper Stickland, Jackson Petty, Richard~Yuanzhe Pang, Julien Dirani, Julian Michael, and Samuel~R. Bowman.
\newblock Gpqa: A graduate-level google-proof q\&a benchmark, 2023.

\bibitem[Sainz et~al.(2023)Sainz, Campos, Garc{\'\i}a-Ferrero, Etxaniz, de~Lacalle, and Agirre]{sainz-etal-2023-nlp}
Oscar Sainz, Jon Campos, Iker Garc{\'\i}a-Ferrero, Julen Etxaniz, Oier~Lopez de~Lacalle, and Eneko Agirre.
\newblock {NLP} evaluation in trouble: On the need to measure {LLM} data contamination for each benchmark.
\newblock In Houda Bouamor, Juan Pino, and Kalika Bali (eds.), \emph{Findings of the Association for Computational Linguistics: EMNLP 2023}, pp.\  10776--10787, Singapore, December 2023. Association for Computational Linguistics.
\newblock \doi{10.18653/v1/2023.findings-emnlp.722}.
\newblock URL \url{https://aclanthology.org/2023.findings-emnlp.722}.

\bibitem[Spearman(1961)]{spearman1961proof}
Charles Spearman.
\newblock The proof and measurement of association between two things.
\newblock \emph{The American Journal of Psychology}, 1961.

\bibitem[Srivastava et~al.(2023)Srivastava, Rastogi, Rao, Shoeb, Abid, Fisch, Brown, Santoro, Gupta, Garriga-Alonso, et~al.]{srivastava2023beyond}
Aarohi Srivastava, Abhinav Rastogi, Abhishek Rao, Abu Awal~Md Shoeb, Abubakar Abid, Adam Fisch, Adam~R Brown, Adam Santoro, Aditya Gupta, Adri{\`a} Garriga-Alonso, et~al.
\newblock Beyond the imitation game: Quantifying and extrapolating the capabilities of language models.
\newblock \emph{Transactions on Machine Learning Research}, 2023.

\bibitem[Team et~al.(2024)Team, Riviere, Pathak, Sessa, Hardin, Bhupatiraju, Hussenot, Mesnard, Shahriari, Ram{\'e}, et~al.]{team2024gemma}
Gemma Team, Morgane Riviere, Shreya Pathak, Pier~Giuseppe Sessa, Cassidy Hardin, Surya Bhupatiraju, L{\'e}onard Hussenot, Thomas Mesnard, Bobak Shahriari, Alexandre Ram{\'e}, et~al.
\newblock Gemma 2: Improving open language models at a practical size.
\newblock \emph{arXiv preprint arXiv:2408.00118}, 2024.

\bibitem[team(2024)]{mistral2024large}
Mistral~AI team.
\newblock Au large.
\newblock \url{https://mistral.ai/news/mistral-large/}, 2024.
\newblock (Accessed on 06/05/2024).

\bibitem[Team(2024{\natexlab{a}})]{qwen1.5}
Qwen Team.
\newblock Introducing qwen1.5, February 2024{\natexlab{a}}.
\newblock URL \url{https://qwenlm.github.io/blog/qwen1.5/}.

\bibitem[Team(2024{\natexlab{b}})]{dbrx2024instruct}
The Mosaic~Research Team.
\newblock Introducing dbrx: A new state-of-the-art open llm.
\newblock \url{https://www.databricks.com/blog/introducing-dbrx-new-state-art-open-llm/}, 2024{\natexlab{b}}.
\newblock (Accessed on 06/05/2024).

\bibitem[Touvron et~al.(2023)Touvron, Martin, Stone, Albert, Almahairi, Babaei, Bashlykov, Batra, Bhargava, Bhosale, et~al.]{touvron2023llama2}
Hugo Touvron, Louis Martin, Kevin Stone, Peter Albert, Amjad Almahairi, Yasmine Babaei, Nikolay Bashlykov, Soumya Batra, Prajjwal Bhargava, Shruti Bhosale, et~al.
\newblock Llama 2: Open foundation and fine-tuned chat models.
\newblock \emph{arXiv preprint arXiv:2307.09288}, 2023.

\bibitem[Wang et~al.(2019)Wang, Singh, Michael, Hill, Levy, and Bowman]{wang2019glue}
Alex Wang, Amanpreet Singh, Julian Michael, Felix Hill, Omer Levy, and Samuel~R. Bowman.
\newblock {GLUE}: A multi-task benchmark and analysis platform for natural language understanding.
\newblock In \emph{ICLR}, 2019.

\bibitem[Wang et~al.(2023)Wang, Kordi, Mishra, Liu, Smith, Khashabi, and Hajishirzi]{wang-etal-2023-self-instruct}
Yizhong Wang, Yeganeh Kordi, Swaroop Mishra, Alisa Liu, Noah~A. Smith, Daniel Khashabi, and Hannaneh Hajishirzi.
\newblock Self-instruct: Aligning language models with self-generated instructions.
\newblock In Anna Rogers, Jordan Boyd-Graber, and Naoaki Okazaki (eds.), \emph{Proceedings of the 61st Annual Meeting of the Association for Computational Linguistics (Volume 1: Long Papers)}, pp.\  13484--13508, Toronto, Canada, July 2023. Association for Computational Linguistics.
\newblock \doi{10.18653/v1/2023.acl-long.754}.
\newblock URL \url{https://aclanthology.org/2023.acl-long.754}.

\bibitem[Wei et~al.(2023)Wei, Wang, Schuurmans, Bosma, Ichter, Xia, Chi, Le, and Zhou]{wei2023chainofthought}
Jason Wei, Xuezhi Wang, Dale Schuurmans, Maarten Bosma, Brian Ichter, Fei Xia, Ed~Chi, Quoc Le, and Denny Zhou.
\newblock Chain-of-thought prompting elicits reasoning in large language models, 2023.

\bibitem[White et~al.(2024)White, Dooley, Roberts, Pal, Feuer, Jain, Shwartz-Ziv, Jain, Saifullah, Naidu, et~al.]{white2024livebench}
Colin White, Samuel Dooley, Manley Roberts, Arka Pal, Ben Feuer, Siddhartha Jain, Ravid Shwartz-Ziv, Neel Jain, Khalid Saifullah, Siddartha Naidu, et~al.
\newblock Livebench: A challenging, contamination-free llm benchmark.
\newblock \emph{arXiv preprint arXiv:2406.19314}, 2024.

\bibitem[Yang et~al.(2024)Yang, Yang, Hui, Zheng, Yu, Zhou, Li, Li, Liu, Huang, et~al.]{yang2024qwen2}
An~Yang, Baosong Yang, Binyuan Hui, Bo~Zheng, Bowen Yu, Chang Zhou, Chengpeng Li, Chengyuan Li, Dayiheng Liu, Fei Huang, et~al.
\newblock Qwen2 technical report.
\newblock \emph{arXiv preprint arXiv:2407.10671}, 2024.

\bibitem[Yang et~al.(2023)Yang, Chiang, Zheng, Gonzalez, and Stoica]{yang2023rethinking}
Shuo Yang, Wei-Lin Chiang, Lianmin Zheng, Joseph~E. Gonzalez, and Ion Stoica.
\newblock Rethinking benchmark and contamination for language models with rephrased samples, 2023.

\bibitem[Zellers et~al.(2019)Zellers, Holtzman, Bisk, Farhadi, and Choi]{zellers2019hellaswag}
Rowan Zellers, Ari Holtzman, Yonatan Bisk, Ali Farhadi, and Yejin Choi.
\newblock Hellaswag: Can a machine really finish your sentence?
\newblock In \emph{Proceedings of the 57th Annual Meeting of the Association for Computational Linguistics}, pp.\  4791--4800, 2019.

\bibitem[Zhao et~al.(2024)Zhao, Ren, Hessel, Cardie, Choi, and Deng]{zhao2024wildchat}
Wenting Zhao, Xiang Ren, Jack Hessel, Claire Cardie, Yejin Choi, and Yuntian Deng.
\newblock Wildchat: 1m chatgpt interaction logs in the wild.
\newblock \emph{International Conference on Learning Representations}, 2024.

\bibitem[Zheng et~al.(2023{\natexlab{a}})Zheng, Chiang, Sheng, Zhuang, Wu, Zhuang, Lin, Li, Li, Xing, Zhang, Gonzalez, and Stoica]{zheng2023judging}
Lianmin Zheng, Wei-Lin Chiang, Ying Sheng, Siyuan Zhuang, Zhanghao Wu, Yonghao Zhuang, Zi~Lin, Zhuohan Li, Dacheng Li, Eric Xing, Hao Zhang, Joseph~E. Gonzalez, and Ion Stoica.
\newblock Judging {LLM}-as-a-judge with {MT}-bench and chatbot arena.
\newblock In \emph{Thirty-seventh Conference on Neural Information Processing Systems Datasets and Benchmarks Track}, 2023{\natexlab{a}}.
\newblock URL \url{https://openreview.net/forum?id=uccHPGDlao}.

\bibitem[Zheng et~al.(2023{\natexlab{b}})Zheng, Chiang, Sheng, Zhuang, Wu, Zhuang, Lin, Li, Li, Xing, Zhang, Gonzalez, and Stoica]{llmjudge}
Lianmin Zheng, Wei-Lin Chiang, Ying Sheng, Siyuan Zhuang, Zhanghao Wu, Yonghao Zhuang, Zi~Lin, Zhuohan Li, Dacheng Li, Eric~P. Xing, Hao Zhang, Joseph~E. Gonzalez, and Ion Stoica.
\newblock Judging llm-as-a-judge with mt-bench and chatbot arena.
\newblock \emph{NeurIPS}, 2023{\natexlab{b}}.

\bibitem[Zhong et~al.(2023)Zhong, Cui, Guo, Liang, Lu, Wang, Saied, Chen, and Duan]{zhong2023agieval}
Wanjun Zhong, Ruixiang Cui, Yiduo Guo, Yaobo Liang, Shuai Lu, Yanlin Wang, Amin Saied, Weizhu Chen, and Nan Duan.
\newblock Agieval: A human-centric benchmark for evaluating foundation models.
\newblock \emph{arXiv preprint arXiv:2304.06364}, 2023.

\bibitem[Zhou et~al.(2023)Zhou, Lu, Mishra, Brahma, Basu, Luan, Zhou, and Hou]{zhou2023instruction}
Jeffrey Zhou, Tianjian Lu, Swaroop Mishra, Siddhartha Brahma, Sujoy Basu, Yi~Luan, Denny Zhou, and Le~Hou.
\newblock Instruction-following evaluation for large language models.
\newblock \emph{arXiv preprint arXiv:2311.07911}, 2023.

\bibitem[Zhu et~al.(2023)Zhu, Frick, Wu, Zhu, and Jiao]{starling2023}
Banghua Zhu, Evan Frick, Tianhao Wu, Hanlin Zhu, and Jiantao Jiao.
\newblock Starling-7b: Improving llm helpfulness \& harmlessness with rlaif, November 2023.

\bibitem[Zhuo et~al.(2024)Zhuo, Vu, Chim, Hu, Yu, Widyasari, Yusuf, Zhan, He, Paul, Brunner, Gong, Hoang, Zebaze, Hong, Li, Kaddour, Xu, Zhang, Yadav, Jain, Gu, Cheng, Liu, Liu, Wang, Lo, Hui, Muennighoff, Fried, Du, de~Vries, and Werra]{zhuo2024bigcodebenchbenchmarkingcodegeneration}
Terry~Yue Zhuo, Minh~Chien Vu, Jenny Chim, Han Hu, Wenhao Yu, Ratnadira Widyasari, Imam Nur~Bani Yusuf, Haolan Zhan, Junda He, Indraneil Paul, Simon Brunner, Chen Gong, Thong Hoang, Armel~Randy Zebaze, Xiaoheng Hong, Wen-Ding Li, Jean Kaddour, Ming Xu, Zhihan Zhang, Prateek Yadav, Naman Jain, Alex Gu, Zhoujun Cheng, Jiawei Liu, Qian Liu, Zijian Wang, David Lo, Binyuan Hui, Niklas Muennighoff, Daniel Fried, Xiaoning Du, Harm de~Vries, and Leandro~Von Werra.
\newblock Bigcodebench: Benchmarking code generation with diverse function calls and complex instructions, 2024.
\newblock URL \url{https://arxiv.org/abs/2406.15877}.

\end{thebibliography}
\bibliographystyle{iclr2025_conference}

\newpage
\appendix
\section{Appendix}
\subsection{Pair Rank Brier Score}
\label{appendix:A.1}
Bootstrapping is a well-established statistical technique for estimating the distribution of an estimator by sampling with replacement from the original dataset. This approach has become increasingly popular for constructing confidence intervals in LLM leaderboards, such as Chatbot Arena~\citep{chiang2024chatbot}. In our proposed evaluation metrics in \autoref{sec:measurement}, such as Separability and Agreement with Confidence Interval, a reliable confidence interval estimation is essential for assessing the performance stability of different models on a given benchmark. Moreover, for metrics like the Pairwise Rank Brier Score, estimating the probability distribution of rank-based model performance is critical. Therefore, applying bootstrapping to the given benchmark provides a straightforward and robust solution for these tasks.

Consider a benchmark consisting of a dataset $D = \{ x_1, x_2, \ldots, x_{|D|} \}$ and a scoring function $f$ that measures the performance of $n$ models $\pi_1, \pi_2, \ldots, \pi_n$ on this dataset. Let $D^*$ denote a bootstrap sample of $D$, and let $f(\pi_i, D^*)$ denote the bootstrapped performance score for model $\pi_i$ using the dataset $D^*$. For simplicity, we use $f^*(\pi_i)$ to denote $f(\pi_i, D^*)$.

To use Brier Score~\citep{brier1950verification} for measuring the accuracy of the given benchmark's probabilistic predictions on model performances, we need to compute the forecasted probability that model $\pi_i$ performs lower than $\pi_j$ on the ground truth measurement for every model pair.
\begin{equation}
    \hat{P}(f^*(\pi_i) < f^*(\pi_j))
\end{equation}
The bootstrapped scores $f^*(\pi_i)$ and $f^*(\pi_j)$ follow an empirical distribution that can be approximated using the Central Limit Theorem (CLT). In most cases, the distribution of $f^*(\pi_i)$ converges asymptotically to a normal distribution, which we also observed in our experiments. Formally, $f^*(\pi_i) \sim \mathcal{N}(\mu_i, \sigma_i^2)$, where $\mu_i$ and $\sigma_i^2$ are the bootstrapped mean and variance, respectively. When this normality assumption does not hold, $\hat{P}(f^*(\pi_i) < f^*(\pi_j))$ can still be estimated from the empirical distribution of the bootstrapped scores.

Let $O_{\pi_i \prec \pi_j}$ denote the ground truth outcome for the model pair $(\pi_i, \pi_j)$, where:
\begin{equation}
    O_{\pi_i \prec \pi_j} = \mathds{1}(\pi_i \text{ performs worse than } \pi_j \text{ on the ground truth evaluation metric})
\end{equation}

The Brier Score Loss is then calculated over the benchmark's prediction for each model pair with respect to the ground truth outcome $O$
\begin{equation}
    \frac{1}{N} \sum_{\{i, j\}} (\hat{P}(f^*(\pi_i) < f^*(\pi_j)) - O_{\pi_i \prec \pi_j})^2 
\end{equation}
where $N$ is the number of model pairs.

\newpage
\subsection{Style Control in Model Evaluation}
\label{appendix:A.2}
To mitigate the potential confounding effects of response style on model evaluation, we implemented an enhanced Bradley-Terry regression framework. This method, inspired by recent LLM evaluation technique \citep{dubois2024length}, controls the influence of answer length on judges' preferences. Recently, Chatbot Arena implemented style control~\citep{stylearena2024} to decouple substance from style in their leaderboard. This approach incorporates style-related features, such as answer length, into the regression model, enabling a distinction between a model's intrinsic capabilities and the influence of these potential confounders like answer style. In essence, style control answers the question: \textit{What would the preference be if everyone has the same style?} This distinction is crucial for a more accurate assessment of model performance without biases.

We extend the standard Bradley-Terry model by introducing additional style features. Let $n$ denote the number of pairwise comparison battles and $M$ the number of models. For each battle $i \in [n]$, we define:
\begin{itemize}
    \item $X_i \in \mathbb{R}^M$: $X_{i,m} = 1$ if model $m$ is on the presented first to the judge, $X_{i,m} = -1$ if presented last, and 0 otherwise.
    \item  $Y_i \in {0,1}$: The outcome, where 1 indicates the first model won.
    \item  $Z_i \in \mathbb{R}^S$: A vector of $S$ style features for the comparison.
\end{itemize}

The traditional Bradley-Terry model estimates model strengths $\beta \in \mathbb{R}^M$ through logistic regression:
\begin{equation}
\hat{\beta} = \arg \min_{\beta \in \mathbb{R}^M} \frac{1}{n}\sum\limits_{i=1}^n \text{BCELoss}(\text{sigmoid}(X_i^\top \beta), Y_i)
\end{equation}
Our enhanced model incorporates style coefficients $\gamma \in \mathbb{R}^S$:
\begin{equation}
\hat{\beta}, \hat{\gamma} = \arg \min_{\beta \in \mathbb{R}^M, \gamma \in \mathbb{R}^S} \frac{1}{n}\sum\limits_{i=1}^n \text{BCELoss}(\text{sigmoid}(X_i^\top \beta + Z_i^{\top}\gamma), Y_i)
\end{equation}

where $\text{BCELoss}$ represents the binary cross-entropy loss. We selected the following style features:
\begin{itemize}
    \item Answer token length
    \item Density of markdown headers, markdown bold elements, and markdown lists.
\end{itemize}

For each feature, we compute a normalized difference
\begin{equation}
\text{normalize}\left(\frac{\text{feature}_A - \text{feature}_B}{\text{feature}_A + \text{feature}_B}\right)
\end{equation}
This normalization technique accounts for the relative difference in features between responses. For instance, the token length difference is normalized as
\begin{equation}
\text{normalize}\left(\frac{\text{length}_A - \text{length}_B}{\text{length}_A + \text{length}_B}\right)
\end{equation}
We chose this approach over alternatives like the hyperbolic tangent normalization used in AlpacaEval
\begin{equation}
\tanh\left(\frac{\text{length}_A - \text{length}_B}{\sigma(\text{length}_A - \text{length}_B)}\right)
\end{equation}
Our method better captures proportional differences, especially in cases where absolute differences may be misleading (e.g., 500 vs. 520 tokens compared to 20 vs. 40 tokens).

The resulting $\hat{\beta}$ coefficients represent model strengths controlled for style effects, while $\hat{\gamma}$ quantifies the impact of each style feature on human preferences. To facilitate meaningful comparisons, we normalize the style coefficients. Our analysis revealed that response length was the most influential style factor, with other markdown-related features having secondary effects.

\begin{table}[H]
\centering
\begin{tabular}{lrrrrrr}
\dataset{} (No Modifications) & & & & & \\
\toprule
Model & Score & Token \# & Header (\%) & Bold (\%) & List (\%) \\
\midrule
gemini-1.5-flash-2-detail & 80.0 & 1035 & 0.010 & 1.503 & 1.288 \\
gemini-1.5-flash-2 & 78.6 & 729 & 0.020 & 1.353 & 1.122 \\
gemini-1.5-flash-2-md & 74.5 & 793 & 0.088 & 1.548 & 1.271 \\
gemini-1.5-flash-2-chatty & 68.2 & 808 & 0.005 & 1.236 & 0.986 \\
gemini-1.5-flash-2-no-md & 61.7 & 574 & 0.003 & 0.924 & 0.979 \\
llama-3.1-70b-detail & 53.5 & 834 & 0.025 & 0.961 & 1.470 \\
llama-3.1-70b-md & 44.9 & 601 & 0.257 & 1.776 & 1.695 \\
llama-3.1-70b & 44.5 & 606 & 0.084 & 0.728 & 1.380 \\
llama-3.1-70b-chatty & 44.3 & 623 & 0.011 & 0.679 & 1.173 \\
llama-3.1-70b-no-md & 37.5 & 522 & 0.010 & 0.123 & 0.986 \\
gpt-3.5-turbo-0125-detail & 25.6 & 416 & 0.008 & 0.447 & 1.540 \\
gpt-3.5-turbo-0125 & 23.1 & 323 & 0.012 & 0.284 & 1.272 \\
gpt-3.5-turbo-0125-md & 22.0 & 328 & 0.372 & 0.877 & 1.601 \\
gpt-3.5-turbo-0125-no-md & 18.0 & 269 & 0.012 & 0.182 & 1.149 \\
gpt-3.5-turbo-0125-chatty & 17.1 & 286 & 0.006 & 0.296 & 1.012 \\
\bottomrule
\\
\dataset{} (Style Control) & & & & & \\
\toprule
Model & Score & Token \# & Header (\%) & Bold (\%) & List (\%) \\
\midrule
gemini-1.5-flash-2 & 75.5 & 729 & 0.020 & 1.353 & 1.122 \\
gemini-1.5-flash-2-detail & 71.2 & 1035 & 0.010 & 1.503 & 1.288 \\
gemini-1.5-flash-2-md & 69.3 & 793 & 0.088 & 1.548 & 1.271 \\
gemini-1.5-flash-2-no-md & 62.5 & 574 & 0.003 & 0.924 & 0.979 \\
gemini-1.5-flash-2-chatty & 61.5 & 808 & 0.005 & 1.236 & 0.986 \\
llama-3.1-70b & 41.7 & 606 & 0.084 & 0.728 & 1.380 \\
llama-3.1-70b-no-md & 39.9 & 522 & 0.010 & 0.123 & 0.986 \\
llama-3.1-70b-detail & 39.8 & 834 & 0.025 & 0.961 & 1.470 \\
llama-3.1-70b-chatty & 39.5 & 623 & 0.011 & 0.679 & 1.173 \\
llama-3.1-70b-md & 34.9 & 601 & 0.257 & 1.776 & 1.695 \\
gpt-3.5-turbo-0125 & 33.2 & 323 & 0.012 & 0.284 & 1.272 \\
gpt-3.5-turbo-0125-no-md & 30.4 & 269 & 0.012 & 0.182 & 1.149 \\
gpt-3.5-turbo-0125-detail & 28.9 & 416 & 0.008 & 0.447 & 1.540 \\
gpt-3.5-turbo-0125-md & 27.9 & 328 & 0.372 & 0.877 & 1.601 \\
gpt-3.5-turbo-0125-chatty & 27.3 & 286 & 0.006 & 0.296 & 1.012 \\
\bottomrule
\end{tabular}
\vspace{0.1em}
\caption{Comparison Between \dataset{} with no modification versus applying style control. 
Prompt for detailed:``You are a helpful assistant who thoroughly explains things with as much detail as possible.'', 
prompt for chatty: ``You are a helpful assistant who is chatty.'', 
prompt for md: ``You are a helpful assistant who uses as much markdown as possible.'', and 
prompt for no-md: ``You are a helpful assistant who never uses markdown.'' 
Token represents average number of tokens, header is average markdown header density per token in percentage, bold is average bold markdown element density per token in percentage, and list is average list markdown element per token in percentage.}
\label{tab:comparison_style_control_causal_full}
\end{table}

\begin{table}[H]
\centering
\resizebox{0.8\columnwidth}{!}{
\begin{tabular}{lrrr}
\toprule
Model          & \dataset{} & Random Sample 1 & Random Sample 2 \\
\midrule
Confiderence Agreement     & \textbf{84.2\%} & 57.5\% & 66.1\%\\
Separability   & \textbf{80.5\%} & 74.7\% & 76.3\%\\
Spearman Correlation   & \textbf{94.7\%} & 64.7\% & 72.5\%\\
Brier Score   & \textbf{0.069} & 0.215 & 0.162\\
\bottomrule
\end{tabular}}
\vspace{0.1em}
\caption{We compare \dataset{} with two sets of 500 prompts randomly sampled from 75K Chatbot Arena user queries. We evaluate the set of top-20 models and compare various statistics across. Each prompt is judged only once by positioning the baseline answer first.}
\vspace{-0.5em}
\label{tab:random_baseline_comparison}
\end{table}

\begin{table}[H]
\centering
\begin{tabular}{lrrrr}
\toprule
& \llamaset{} & Random 1 & Random 2 & \dataset{}-500 \\
\midrule
Confidence Agreement & 86.0\% & 55.8\% & 58.1\% & 88.4\% \\
Separability & 84.4\% & 68.9\% & 64.4\% & 88.9\% \\
Spearman Correlation & 96.4\% & 73.3\% & 70.9\% & 96.4\% \\
\bottomrule
\end{tabular}
\vspace{0.1em}
\caption{Comparing \llamaset{} against two random baselines on 10 of the 20 models outlined in the paper. We observe similar improvement in benchmark quality, suggesting \method{} is robust across different choices of LLM annotators.}
\label{tab:comparison_llama_annotator}
\end{table}

\begin{table}[H]
\centering
\begin{tabular}{lr}
\toprule
          & \dataset{} \\
\midrule
Confiderence Agreement     & 98.6\% \\
Spearman Correlation       & 96.7\% \\
Kendall Tau Correlation    & 87.4\% \\
Brier Score                & 0.055 \\
\bottomrule
\end{tabular}
\vspace{0.1em}
\caption{We compare \dataset{} (gpt-4-1106-preview as judge) to Chatbot Arena Category Hard Prompt (English) on the same set of top-20 models.  By comparing \dataset{} to a challenging distribution of queries from Chatbot Arena, we obtain even higher alignment to human preferences.}
\vspace{-0.5em}
\label{tab:comparison_hard_category}
\end{table}

\begin{table}[H]
\centering
\begin{minipage}{.5\textwidth}
\centering
\caption*{OpenAI GPT Series}
\begin{tabular}{lrr|}
\toprule
 & GPT-4-turbo & Ensemble \\
\midrule
gpt-4-turbo & 0 & 0 \\
gpt-4-0314 & 1 & 1 \\
gpt-4-0613 & 0 & -2 \\
gpt-3.5-turbo-0613 & 1 & -1 \\
gpt-3.5-turbo-0314 & 1 & 0 \\
\midrule
column average & 0.6 & -0.4 \\
\bottomrule
\end{tabular}
\end{minipage}%
\begin{minipage}{.5\textwidth}
\centering
\caption*{Anthropic Claude Series}
\begin{tabular}{lrr}
\toprule
 & GPT-4-turbo & Ensemble \\
\midrule
claude-3-opus & 0 & 0 \\
claude-3-sonnet & -1 & -1 \\
claude-2.0 & -2 & 0 \\
claude-2.1 & -1 & 3 \\
\\
\midrule
column average & -0.8 & 0.4 \\
\bottomrule
\end{tabular}
\end{minipage}
\caption{Comparing bias in GPT-4-Turbo as a Judge and Ensemble-as-Judge. We calculate the ranking shift by comparing the human preference ranking (by Chatbot Arena Category Hard Leaderboard) and LLM-judge ranking on OpenAI GPT Series and Anthropic Claude Series. Results show both methods have relatively small shifts, but Ensemble-as-Judge produces a more balanced rank difference than GPT-4-Turbo Judge, suggesting a smaller self-bias than single LLM as a Judge.}
\label{tab:comparison_self_bias}
\end{table}

\begin{table}[H]
\centering
\resizebox{1\columnwidth}{!}{
\begin{tabular}{lrrrrrrr}
\toprule
Quality Score    & 1+ & 2+ & 3+ & 4+ & 5+ & 6+ & 7+\\
\midrule
\% of queries    & 95.4 & 83.5 & 61.9 & 48.7 & 33.8 & 17.9 & 0.2\\
\midrule
\midrule
Qualities        & \small{Specificity} &  \small{Domain-knowledge} & \small{Complexity} & \small{Problem-solving} & \small{Creativity} & \small{Tech. Accuracy} & \small{Real-world}\\
\midrule
\% of queries    & 57.3 &  63.4 & 35.0 & 34.9 & 26.1 & 39.0 & 87.9\\
\bottomrule
\end{tabular}}
\vspace{0.1em}
\caption{First row is the percentage of queries with quality scores of the column or more in 75K Chatbot Arena data assigned by GPT-3.5-Turbo. Second row is the percentage of queries in 75K Chatbot Arena labeled by GPT-3.5-Turbo with each of the 7 qualities.}
\vspace{-0.5em}
\label{tab:criteria_score_percentage}
\end{table}

\begin{table}[H]
\centering
\begin{minipage}{.5\textwidth}
\centering
\caption*{Avg. Token Length}
\begin{tabular}{lrr}
\toprule
 & Pearson & Spearman \\
\midrule
No Modification & 0.364 & 0.125 \\
Style Control & 0.193 & -0.025 \\
\bottomrule
\end{tabular}
\end{minipage}%
\begin{minipage}{.5\textwidth}
\centering
\caption*{Naive Verbose Policy}
\begin{tabular}{lrr}
\toprule
 & Pearson & Spearman \\
\midrule
No Modification & 0.397 & 0.165 \\
Style Control & 0.231 & 0.028 \\
\bottomrule
\end{tabular}
\end{minipage}
\caption{Left: Comparing correlation between model score and average token length between GPT-4-Turbo as Judge with no modification versus style controlled. Right: Comparing correlation to model score produced via a ``verbose policy'', a judge which always picks the longer response. In both cases, style control effectively reduces the correlation to verbosity.}
\label{tab:comparison_style_control_verbosity_correlation}
\end{table}

\begin{table}[H]
\centering
\begin{tabular}{lccc}
\toprule
\textbf{Model Name} & \textbf{Win Rate} & \textbf{CI Interval} & \textbf{Average Token \#} \\
\midrule
Claude-3-5-Sonnet-20240620 & 79.3 & (-2.1, 2.0) & 567 \\
GPT-4O-2024-05-13 & 79.2 & (-1.9, 1.7) & 696 \\
GPT-4-0125-Preview & 78.0 & (-2.1, 2.4) & 619 \\
GPT-4O-2024-08-06 & 77.9 & (-2.0, 2.1) & 594 \\
Athene-70B & 77.6 & (-2.7, 2.2) & 684 \\
GPT-4O-Mini & 74.9 & (-2.5, 1.9) & 668 \\
Gemini-1.5-Pro-API-Preview & 72.0 & (-2.1, 2.5) & 676 \\
Mistral-Large-2407 & 70.4 & (-1.6, 2.1) & 623 \\
LLaMA-3.1-405B-Instruct-FP8 & 69.3 & (-2.4, 2.2) & 658 \\
GLM-4-0520 & 63.8 & (-2.9, 2.8) & 636 \\
Yi-Large & 63.7 & (-2.6, 2.4) & 626 \\
DeepSeek-Coder-V2 & 62.3 & (-2.1, 1.8) & 578 \\
Claude-3-Opus-20240229 & 60.4 & (-2.5, 2.5) & 541 \\
Gemma-2-27B-IT & 57.5 & (-2.1, 2.4) & 577 \\
LLaMA-3.1-70B-Instruct & 55.7 & (-2.9, 2.7) & 628 \\
GLM-4-0116 & 55.7 & (-2.4, 2.3) & 622 \\
GPT-4-0314 & 50.0 & (0.0, 0.0) & 423 \\
Gemini-1.5-Flash-API-Preview & 49.6 & (-2.2, 2.8) & 642 \\
Qwen2-72B-Instruct & 46.9 & (-2.5, 2.7) & 515 \\
Claude-3-Sonnet-20240229 & 46.8 & (-2.3, 2.7) & 552 \\
LLaMA-3-70B-Instruct & 46.6 & (-2.3, 2.6) & 591 \\
Claude-3-Haiku-20240307 & 41.5 & (-2.5, 2.5) & 505 \\
GPT-4-0613 & 37.9 & (-2.8, 2.4) & 354 \\
Mistral-Large-2402 & 37.7 & (-2.1, 2.6) & 400 \\
Mixtral-8x22B-Instruct-V0.1 & 36.4 & (-2.4, 2.6) & 430 \\
Qwen1.5-72B-Chat & 36.1 & (-2.0, 2.7) & 474 \\
Phi-3-Medium-4K-Instruct & 33.4 & (-2.6, 2.1) & 517 \\
Mistral-Medium & 31.9 & (-1.9, 2.2) & 485 \\
InternLM2.5-20B-Chat & 31.2 & (-2.4, 2.8) & 576 \\
Phi-3-Small-8K-Instruct & 29.8 & (-1.8, 1.9) & 568 \\
Mistral-Next & 27.4 & (-2.4, 2.4) & 297 \\
GPT-3.5-Turbo-0613 & 24.8 & (-1.9, 2.3) & 401 \\
DBRX-Instruct-Preview & 24.6 & (-2.0, 2.6) & 415 \\
InternLM2-20B-Chat & 24.4 & (-2.0, 2.2) & 667 \\
Mixtral-8x7B-Instruct-V0.1 & 23.4 & (-2.0, 1.9) & 457 \\
GPT-3.5-Turbo-0125 & 23.3 & (-2.2, 1.9) & 329 \\
Yi-34B-Chat & 23.1 & (-1.6, 1.8) & 611 \\
Starling-LM-7B-Beta & 23.0 & (-1.8, 1.8) & 530 \\
LLaMA-3.1-8B-Instruct & 21.3 & (-1.9, 2.2) & 861 \\
Snorkel-Mistral-PairRM-DPO & 20.7 & (-1.8, 2.2) & 564 \\
LLaMA-3-8B-Instruct & 20.6 & (-2.0, 1.9) & 585 \\
GPT-3.5-Turbo-1106 & 18.9 & (-1.8, 1.6) & 285 \\
Gemini-1.0-Pro & 17.8 & (-1.2, 2.2) & 322 \\
Command-R & 17.0 & (-1.7, 1.8) & 432 \\
Phi-3-Mini-128K-Instruct & 15.4 & (-1.4, 1.4) & 609 \\
Tulu-2-DPO-70B & 15.0 & (-1.6, 1.3) & 550 \\
Starling-LM-7B-Alpha & 12.8 & (-1.6, 1.4) & 483 \\
Gemma-1.1-7B-IT & 12.1 & (-1.3, 1.3) & 341 \\
LLaMA-2-70B-Chat-HF & 11.6 & (-1.5, 1.2) & 595 \\
Vicuna-33B-V1.3 & 8.6 & (-1.1, 1.1) & 451 \\
Gemma-7B-IT & 7.5 & (-1.2, 1.3) & 378 \\
LLaMA-2-7B-Chat-HF & 4.6 & (-0.8, 0.8) & 561 \\
Gemma-1.1-2B-IT & 3.4 & (-0.6, 0.8) & 316 \\
Gemma-2B-IT & 3.0 & (-0.6, 0.6) & 369 \\
\end{tabular}
\vspace{0.2em}
\caption{\dataset{} Leaderboard (baseline: GPT-4-0314) with some additional models~\citep{athene70b2024, deepseekai2024deepseekcoderv2breakingbarrierclosedsource, glm2024chatglm, yang2024qwen2, cai2024internlm2, abdin2024phi, team2024gemma}.}
\end{table}

\begin{figure}[H]
    \centering
    \includegraphics[width=1.0\textwidth]{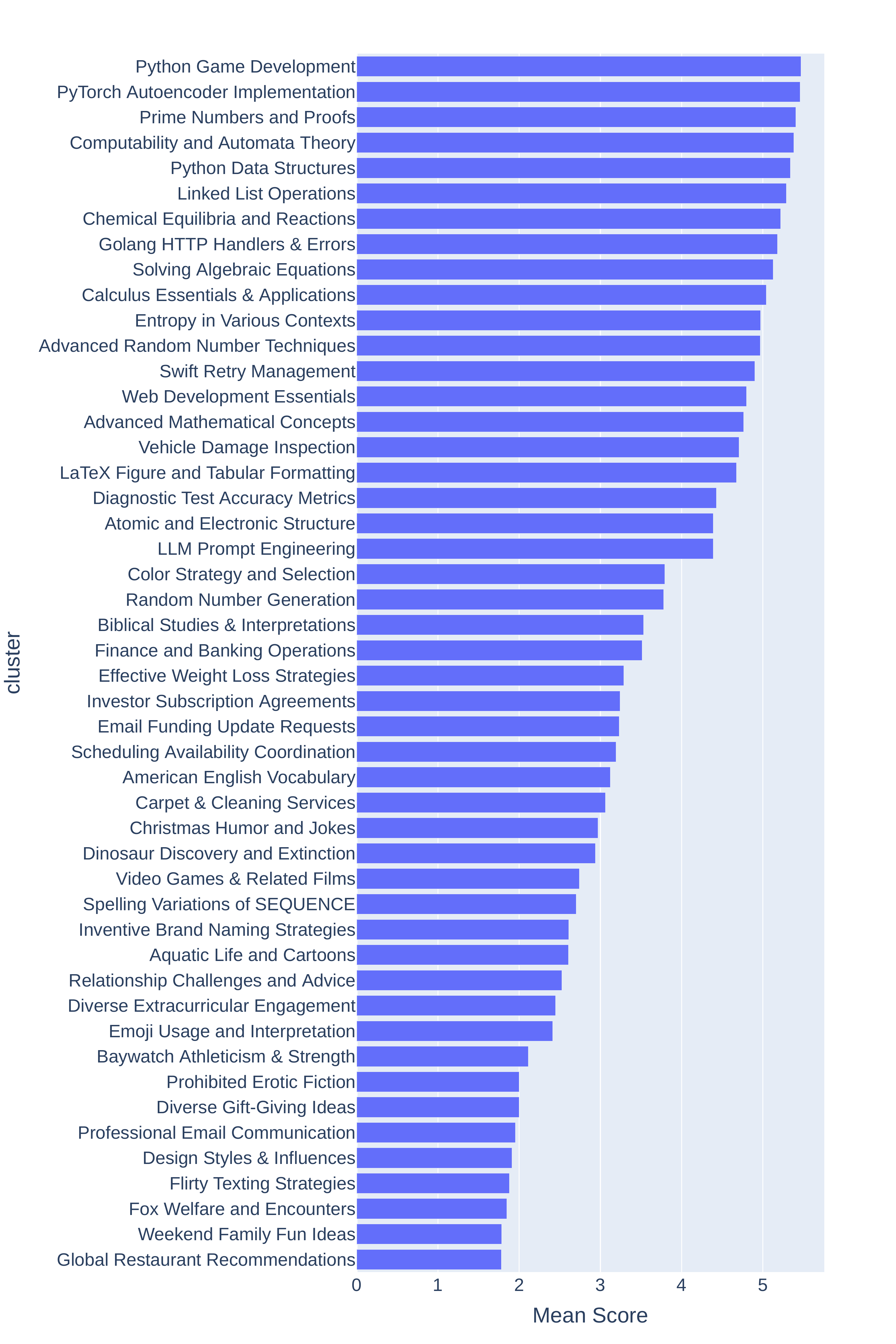}
    \caption{A more complete selection of mean scores of various topic clusters in descending order.}
    \label{fig:cluster_distribution_full}
\end{figure}

\section{Examples}
\label{sec:examples}
\lstset{
basicstyle=\small\ttfamily,
columns=fullflexible,
breaklines=true,
mathescape,
breakindent=0pt,
}
\begin{example}
\small
\textbf{Cluster 1: Greetings and Well-Being Inquiry} (Mean Score: 2.7)
\begin{lstlisting}
Yo, what up my brother (Qualities: None)
\end{lstlisting}

\textbf{Cluster 2: US Presidents Query} (Mean Score: 3.2)
\begin{lstlisting}
Who was the president of the US in 1975 (Qualities: Specificity, Domain-Knowledge, Technical Accuracy, Real-World)
\end{lstlisting}

\textbf{Cluster 3: Physics Problem Solving} (Mean Score: 5.0)
\begin{lstlisting}
A 50,000 kg airplane initially flying at a speed of 60.0 m/s accelerates at 5.0 m/s2 for 600 meters. What is its velocity after this acceleration? What is the net force that caused this acceleration? (Qualities: Specificity, Domain-Knowledge, Complexity, Problem-Solving, Technical Accuracy, Real-World) 
\end{lstlisting}

\textbf{Cluster 4: OpenCV Image Processing Technique} (Mean Score: 5.5)
\begin{lstlisting}
you are given a task to detect number of faces in each frame of any video using pytorch and display the number in the final edited video. (Qualities: All)
\end{lstlisting}
\end{example}

\newpage
\section{Prompts}
\label{sec:prompts}

\begin{userinput}{Prompt Quality Systems Instruction:}
Your task is to evaluate how well the following input prompts can assess the capabilities of advanced AI assistants. For the input prompt, please analyze it based on the following 7 criteria. For each criteria, make sure to explain before determine whether the input satisfy it.
 \vspace{1em}
 
1. Specificity: Does the prompt ask for a specific, well-defined output without leaving any ambiguity? This allows the AI to demonstrate its ability to follow instructions and generate a precise, targeted response. \\
2. Domain Knowledge: Does the prompt test the AI's knowledge and understanding in a specific domain or set of domains? The prompt must demand the AI to have a strong prior knowledge or mastery of domain-specific concepts, theories, or principles. \\
3. Complexity: Does the prompt have multiple components, variables, or levels of depth and nuance? This assesses the AI's capability to handle complex, multi-faceted problems beyond simple queries. \\
4. Problem-Solving: Does the prompt require active problem-solving: analyzing and clearly defining the problem and systematically devising and implementing a solution? Note active problem-solving is not simply reciting facts or following a fixed set of instructions. \\
5. Creativity: Does the prompt require a creative approach or solution? This tests the AI’s ability to generate novel ideas tailored to the specific needs of the request or problem at hand. \\
6. Technical Accuracy: Does the prompt require an answer with a high degree of technical accuracy, correctness and precision? This assesses the reliability and truthfulness of the AI's outputs. \\
7. Real-World Application: Does the prompt relate to real-world applications? This tests the AI's ability to provide practical and actionable information that could be implemented in real-life scenarios.

 \vspace{1em}
After analyzing the input prompt based on these criteria, you must list the criteria numbers that the prompt satisfies in the format of a Python array. For example, "Criteria Satisfied: [1, 2, 4, 6, 7]".
\end{userinput}

\begin{userinput}{LLM-Judge System Instruction}
    Please act as an impartial judge and evaluate the quality of the responses provided by two AI assistants to the user prompt displayed below. You will be given assistant A's answer and assistant B's answer. Your job is to evaluate which assistant's answer is better. \vspace{1em}
    
    Begin your evaluation by generating your own answer to the prompt. You must provide your answers before judging any answers. \vspace{1em}
    
    When evaluating the assistants' answers, compare both assistants' answers with your answer. You must identify and correct any mistakes or inaccurate information. \vspace{1em}
    
    Then consider if the assistant's answers are helpful, relevant, and concise. Helpful means the answer correctly responds to the prompt or follows the instructions. Note when user prompt has any ambiguity or more than one interpretation, it is more helpful and appropriate to ask for clarifications or more information from the user than providing an answer based on assumptions. Relevant means all parts of the response closely connect or are appropriate to what is being asked. Concise means the response is clear and not verbose or excessive. \vspace{1em}
    
    Then consider the creativity and novelty of the assistant's answers when needed. Finally, identify any missing important information in the assistants' answers that would be beneficial to include when responding to the user prompt.
    \vspace{1em}
    
    After providing your explanation, you must output only one of the following choices as your final verdict with a label:

    \begin{enumerate}
        \item Assistant A is significantly better: [[A>{}>B]]
        \item Assistant A is slightly better: [[A>B]]
        \item Tie, relatively the same: [[A=B]]
        \item Assistant B is slightly better: [[B>A]]
        \item Assistant B is significantly better: [[B>{}>A]
    \end{enumerate}
     \vspace{1em}
    
    Example output: "My final verdict is tie: [[A=B]]".
\end{userinput}

\newpage
\section{Additional Examples}
\label{sec:additional_examples}

\begin{example}
\small
\textbf{User:} Use ABC notation to write a melody in the style of a folk tune.

\vspace{1em}

\textbf{GPT-4:}

\begin{verbatim}
X:1 T:Untitled Folk Tune M:4/4 L:1/8 
K:G|:G2A2|B2A2|G2E2|D4|E2F2|G2F2|E2C2|B,4|A2B2|c2B2|A2F2|E4|D2E2|F2E2
|D2B,2|C4:|
\end{verbatim}

\textbf{GPT-3.5-Turbo:}
Sure! Here’s a simple melody in the style of a folk tune written in ABC notation:

\begin{verbatim}
X:1 T:Folk Tune M:4/4 L:1/4 K:C 
CDE|FGA|BAG|FED| CDE|FGA|BAG|c2z| CDE|FGA|BAG|FED| CDE|FGA|BAG|c2z||
\end{verbatim}

This melody is in the key of C major and consists of four measures. Feel free to modify it or add more sections to create your own folk tune!

\vspace{1em}

\textbf{Judgement:} My own answer to the prompt would be:

\begin{verbatim}
    X:1 T:My Folk Melody M:3/4 L:1/8 
    K:D A2|:d2f2a2|b2a2f2|e2d2B2|A4F2| G2A2B2|A2G2F2|E2D2F2|1D6:|2D4A2| 
    |:d2f2a2|b2a2f2|e2c2A2|B4A2| G2A2B2|A2G2F2|E2D2F2|1D6:|2D4||
\end{verbatim}

This melody is in the key of D major, has a 3/4 time signature, and is structured in a typical AAB form common in folk music. It uses a range of an octave, which is typical for folk tunes that need to be easily singable.

Now, let’s evaluate the responses from Assistant A and Assistant B.

Assistant A’s Answer: Assistant A provides a melody in ABC notation that is coherent and follows the structure of a folk tune. The melody is in 4/4 time, which is common for folk music, and it is in the key of G major. The melody has a two-part form (indicated by the repeat signs |: and :|), which is typical in folk music. The range of the melody is appropriate for a folk tune, and the rhythm is simple and repetitive, which is characteristic of the genre.

Assistant B’s Answer: Assistant B also provides a melody in ABC notation that is coherent and could be considered in the style of a folk tune. The melody is in 4/4 time and the key of C major. The melody consists of four measures repeated twice, which is a simple structure that could be found in folk music. However, Assistant B’s melody is less traditional in its form compared to Assistant A’s, as it does not have a distinct A and B section, which is more common in folk tunes. Additionally, the use of “z” for a half-measure rest is less typical in folk tunes, which usually have a more continuous melodic line.

Comparison: Both assistants provided melodies that could be considered folk tunes. Assistant A’s melody has a more traditional structure with a clear A and B section, while Assistant B’s melody is simpler and does not follow the traditional AAB or AB form. Assistant A’s melody is also more detailed in terms of the ABC notation, providing a full eight-measure section that is repeated, which gives a more complete sense of the tune. Assistant B’s melody is correct in its notation but is more basic and less representative of a traditional folk tune structure.

In terms of helpfulness, relevance, and conciseness, both assistants provided relevant and concise answers. However, Assistant A’s answer is slightly more helpful due to the more traditional structure of the provided melody.

Creativity and novelty are not particularly applicable in this context, as the task is to create a melody in a traditional style rather than to innovate. Both assistants showed an understanding of the folk tune style, but Assistant A’s answer aligns more closely with traditional folk structures.

There are no significant mistakes or inaccurate information in either assistant’s answers. However, Assistant B’s use of the rest “z” in a folk tune is less conventional.

My final verdict is: Assistant A is slightly better: [[A>B]].
\end{example}









\end{document}